\documentclass[times, twoside, watermark]{BioRxiv}
\usepackage{multirow}
\usepackage{graphicx}
\usepackage{amsmath}
\usepackage{amssymb}
\usepackage{afterpage}
\usepackage{xcolor}
\usepackage{dblfloatfix}
\usepackage{array,multirow,adjustbox,ragged2e,xcolor,graphicx,fontawesome5}
\usepackage{tikz}
\usepackage{graphicx}
\usepackage{xcolor}

\definecolor{PDSeqColor}{HTML}{1B9E77}
\definecolor{PDStrColor}{HTML}{1F6FEB}
\definecolor{PDFnColor}{HTML}{D95F02}
\definecolor{PDTxtColor}{HTML}{8A3FFC}
\definecolor{PDPythonColor}{HTML}{3776AB}

\newcommand{\PDcode}[1]{\href{#1}{\textcolor{PDPythonColor}{\faPython}}}
\newcommand{\PDref}[1]{\cite{#1}}
\newcolumntype{D}[1]{>{\Centering\arraybackslash}m{#1}}
\newcolumntype{E}[1]{>{\RaggedRight\arraybackslash}m{#1}}

\newcommand{\PDgroup}[2]{\multirow{#1}{=}{\centering\bfseries\shortstack{#2}}}
\newcommand{\PDgh}[1]{\PDcode{https://github.com/#1}}

\newcommand{\PDrow}[7]{& #1 & \PDref{#2} & #3 &
  \csname PDm#4\endcsname & #5 & #6 & #7 \\}
\newcommand{\PDseparator}{\noalign{\vskip .7pt}\hline\noalign{\vskip 1.3pt}}
\newcommand{\PDheader}{\hline
\rule{0pt}{3.3ex}\textbf{Design task} & \textbf{Method} & \textbf{Ref.} &
\textbf{Generation approach} & \multicolumn{4}{c}{\textbf{Modality}} &
\textbf{Architecture} & \textbf{Conditioning} & \textbf{Impl.} \\
 & & & & Seq & Str & Fn & Txt & & & \\[.5pt]
\hline}

\DeclareCaptionLabelFormat{appendixS}{#1~S#2}

\begin{document}

\renewcommand\Affilfont{\normalfont\small\raggedright}
\makeatletter
\renewcommand\Affilfont{%
  \normalfont\small\raggedright
  \linespread{1.25}\selectfont
}
\makeatother

\title{Generative machine learning for protein design}

\author[1,+]{Senura Hansaja Wanasekara}
\author[2,3,+]{Manoj M Wagle}
\author[4]{Minh-Duong Nguyen}
\author[1]{Xiaochen Liu}
\author[5]{Nguyen H. Tran}
\author[2,6*]{Manolis Kellis}
\author[1*]{Ken-Tye Yong}

\affil[1]{School of Biomedical Engineering, Faculty of Engineering, The University of Sydney, Darlington, 2008 NSW, Australia}
\affil[2]{Computer Science and Artificial Intelligence Laboratory, Massachusetts Institute of Technology, Cambridge, 02139 Massachusetts, USA}
\affil[3]{School of Mathematics and Statistics, Faculty of Science, The University of Sydney, Camperdown, 2050 NSW, Australia}
\affil[4]{College of Engineering and Computer Science, VinUniversity, Hanoi, Vietnam}
\affil[5]{School of Computer Science, Faculty of Engineering, The University of Sydney, Camperdown, 2006 NSW, Australia}
\affil[6]{The Broad Institute of MIT and Harvard, 415 Main Street, Cambridge, 02142 Massachusetts, USA}
\affil[+]{Co-first authors}
\affil[*]{Correspondence: [ken.yong@sydney.edu.au], [manoli@mit.edu]}

\maketitle

\section*{Abstract}
\begin{abstract}

Designing proteins with specific biological functions is of considerable interest for both the development of new therapeutics and the engineering of enzymes, yet achieving the intended activity remains challenging. Generative artificial intelligence (AI) has emerged as a powerful tool to explore the vast space of possible designs by learning from protein sequence and structural data. Here, we review 59 generative approaches across six protein design tasks, including sequence generation, backbone generation, structure-conditioned sequence design, joint sequence-structure design, motif scaffolding, and binder and antibody design. We outline common generative approaches and protein representations and discuss how structural constraints and functional information are utilized during generation. This is followed by a discussion of the computational and experimental approaches used to evaluate designed proteins and what such assessments reveal about their properties. Finally, we discuss current limitations in designing proteins beyond well-represented families and incorporating experimental feedback into the design process. We expect that rapid advances in generative modeling will enable more reliable design of a wider range of protein structures and functions over the coming years.

\end{abstract}

\section*{Introduction}\hfill

\begin{figure*}[!htbp]
    \centering
    \includegraphics[width=\dimexpr\textwidth-7em\relax]{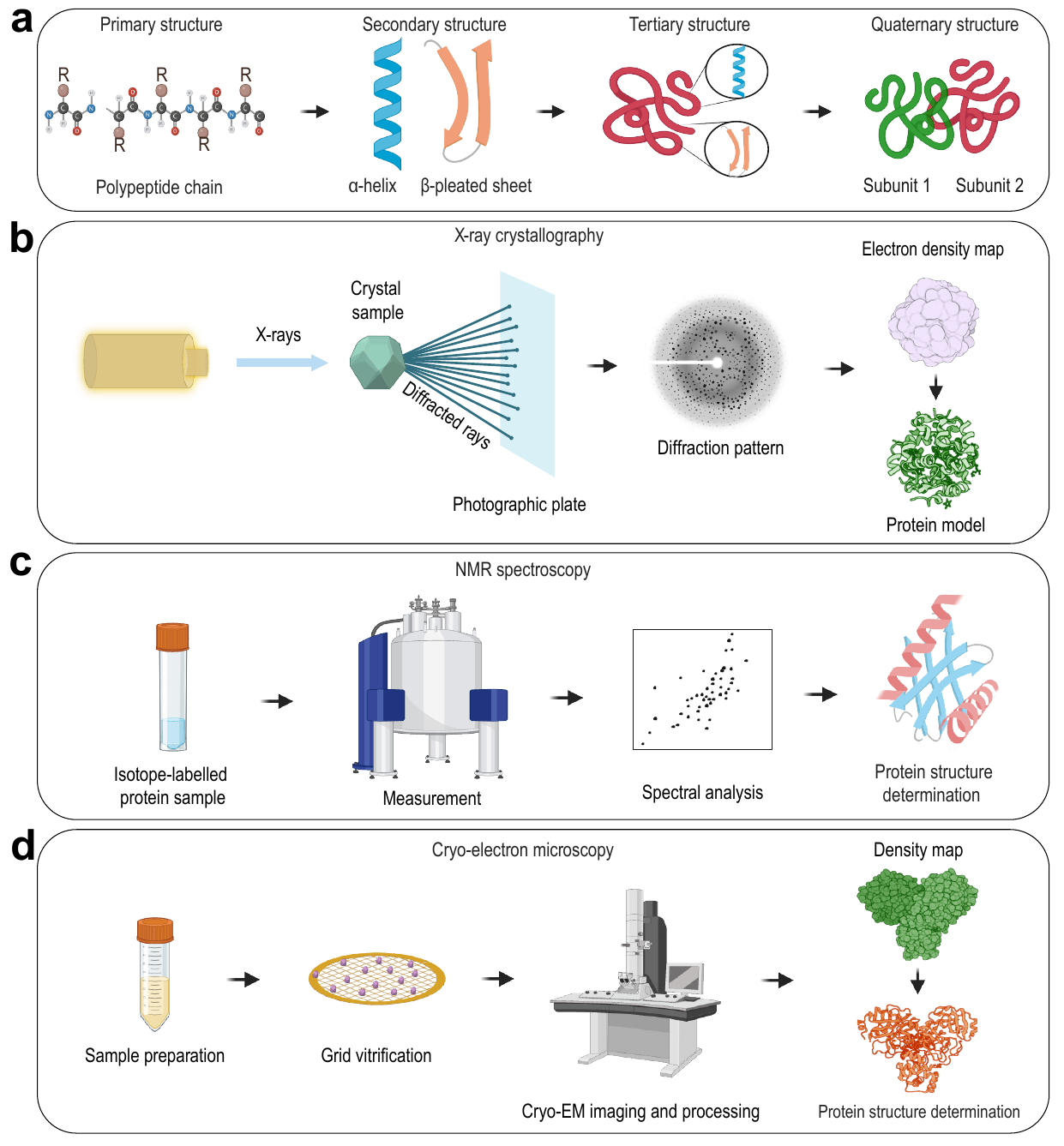}
    \caption{\textbf{Overview of protein structural organization and experimental structure determination}
    (a) Schematic illustration of primary, secondary, tertiary, and quaternary protein structure, showing the amino acid chain, local secondary structure elements (α-helix and β-pleated sheet), three-dimensional folding of a polypeptide, and assembly of multiple subunits, respectively.
    (b) X-ray crystallography workflow, where diffraction measurements from a protein crystal are used to reconstruct an electron density map for building and refining an atomic model of the protein.
    (c) Nuclear magnetic resonance (NMR) spectroscopy showing measurements from an isotope-labeled protein sample and spectral analysis for determining protein structure.
    (d) Cryo-electron microscopy (Cryo-EM) showing sample preparation, grid vitrification, imaging, and computational reconstruction of a three-dimensional map for building and refining a structural model.}\label{fig1}
\end{figure*}

Proteins are complex biomolecules comprised of amino acids and are essential for carrying out diverse cellular functions, including structural support, metabolism, chemical and immune regulation. Their function is closely governed by the interplay of sequence, structure, and conformational dynamics, and the ability to design these molecules has become a cornerstone of modern biology \cite{alberts2015essential, anfinsen1973principles, huang2016coming} (Fig. \ref{fig1}a). This has facilitated the design of therapeutics, including vaccines, antibodies, drug-delivery vehicles, and engineered enzymes for specific applications \cite{kuhlman2019advances, bennett2026atomically}.  However, the relationships between sequence, structure, and function are complex. A protein can adopt multiple conformational states, and designing for a particular activity may require accounting for more than one static structure. The sheer range of possible sequences and conformations makes it very difficult to determine which designs will achieve the intended structural or functional properties.\\

Experimental and computational approaches provide complementary information for understanding and designing proteins. Experimental techniques such as X-ray crystallography, NMR, and Cryo-EM underpin the data deposited in resources such as the Protein Data Bank (PDB) \cite{berman2002protein}. Molecular dynamics simulations complement these measurements by examining protein motions, while improved sampling methods help explore conformational states that may be difficult to access in conventional simulations. Nevertheless, their usefulness depends on force-field accuracy and sufficient sampling, which can require substantial computational resources \cite{hollingsworth2018molecular,karplus2002molecular}. Traditional computational design frameworks such as Rosetta use structural modeling and energy calculations to search for sequences compatible with a specified design objective \cite{kuhlman2019advances,leaver2011rosetta3}. The design and experimental characterization of Top7 demonstrated that these approaches could produce a protein with a novel fold \cite{kuhlman2003design}. However, identifying successful designs can require extensive computational searches followed by experimental testing and further refinement \cite{kuhlman2019advances,leaver2011rosetta3}.\\

\begin{figure*}[t]
\centering
\begingroup

\definecolor{TimelineOrange}{HTML}{FF913B}
\definecolor{TimelineBlue}{HTML}{ECF8FA}
\definecolor{TimelineLine}{HTML}{808080}

\newcommand{\TimelineFont}{%
  \fontfamily{phv}\selectfont
  \fontsize{8.5}{10.5}\selectfont
}


\newcommand{\TimelineLeft}[4]{%
  \draw[TimelineLine, line width=0.35pt]
    (-0.50,#1) -- (-1.60,#1);
  \node[year] at (0,#1) {#2};
  \node[
    milestone,
    anchor=east,
    align=flush right,
    text width=#3
  ] at (-1.60,#1) {#4};
}

\newcommand{\TimelineRight}[4]{%
  \draw[TimelineLine, line width=0.35pt]
    (0.50,#1) -- (1.60,#1);
  \node[year] at (0,#1) {#2};
  \node[
    milestone,
    anchor=west,
    align=flush left,
    text width=#3
  ] at (1.60,#1) {#4};
}

\resizebox{\dimexpr\textwidth-7em\relax}{!}{%
\begin{tikzpicture}[
  x=1cm,
  y=-1cm,
  milestone/.style={
    fill=TimelineBlue,
    draw=none,
    text=black,
    font=\TimelineFont,
    inner xsep=5pt,
    inner ysep=5pt,
    outer sep=0pt
  },
  year/.style={
    text=white,
    font=\TimelineFont,
    inner sep=0pt,
    outer sep=0pt
  }
]

\path[use as bounding box]
  (-7.85,-0.55) rectangle (7.85,17.15);

\path[fill=TimelineOrange, draw=none]
  (-0.50,-0.50) --
  ( 0.50,-0.50) --
  ( 0.50,16.25) --
  ( 0.82,16.25) --
  ( 0.00,17.15) --
  (-0.82,16.25) --
  (-0.50,16.25) -- cycle;

\node[
  anchor=north west,
  font=\sffamily\bfseries\fontsize{20}{22}\selectfont,
  inner sep=0pt
] at (-7.65,-0.45) {a};

\TimelineLeft{0.00}{1958}{4.70cm}{%
  The first protein structure, myoglobin, was
  determined by X-ray crystallography~\cite{kendrew1958three}
}

\TimelineRight{0.98}{1961}{5.80cm}{%
  Ribonuclease refolding experiments supported the
  role of amino acid sequence in determining the
  native protein structure~\cite{anfinsen1961kinetics}
}

\TimelineLeft{1.96}{1971}{3.85cm}{%
  The Protein data bank (PDB) was
  established~\cite{berman2002protein}
}

\TimelineRight{2.94}{1977}{5.10cm}{%
  The first molecular dynamics simulation of a
  protein was performed~\cite{mccammon1977dynamics}
}

\TimelineLeft{3.92}{1985}{5.10cm}{%
  The first solution-state protein structure was
  determined using NMR spectroscopy~\cite{williamson1985solution}
}

\TimelineRight{4.90}{1999}{5.55cm}{%
  Replica-exchange molecular dynamics was
  introduced for enhanced conformational
  sampling~\cite{sugita1999replica}
}

\TimelineLeft{5.88}{2003}{5.20cm}{%
  Computational design of Top7 demonstrated a
  protein with a novel fold~\cite{kuhlman2003design}
}

\TimelineRight{6.86}{2013}{5.10cm}{%
  Direct electron detectors enabled near-atomic
  cryo-EM of large complexes~\cite{bai2013ribosome}
}

\TimelineLeft{7.84}{2018}{6.05cm}{%
  AlphaFold achieved leading performance in protein
  structure prediction at CASP13~\cite{senior2020improved}
}

\TimelineRight{8.82}{2020}{5.70cm}{%
  AlphaFold2 achieved near-experimental accuracy
  for many targets at CASP14~\cite{jumper2021highly}
}

\TimelineLeft{9.80}{2021}{6.05cm}{%
  RoseTTAFold introduced a three-track network for
  protein structure prediction~\cite{baek2021accurate}
}

\TimelineRight{10.78}{2022}{5.55cm}{%
  ProteinMPNN advanced structure-conditioned
  protein sequence design~\cite{dauparas2022robust}
}

\TimelineLeft{11.76}{2023}{6.05cm}{%
  The RFdiffusion study demonstrated backbone generation
  and constrained protein design across diverse
  tasks~\cite{watson2023novo}
}

\TimelineRight{12.74}{2024}{5.50cm}{%
  AlphaFold3 extended structure prediction to full
  biomolecular complexes~\cite{abramson2024accurate}
}

\TimelineLeft{13.72}{2024}{5.45cm}{%
  ESM3 jointly modeled protein sequence,
  structure, and function within a single
  framework~\cite{hayes2025simulating}
}

\TimelineRight{14.70}{2025}{5.20cm}{%
  RFdiffusion2 enabled enzyme scaffolding
  from specified catalytic atom
  arrangements~\cite{ahern2026atom}
}

\TimelineLeft{15.68}{2026}{6.05cm}{%
  RFdiffusion3 demonstrated atomic functional-site
  scaffolding with experimentally active designed
  enzymes~\cite{butcher2026atomic}
}

\end{tikzpicture}%
}

\endgroup

\caption{%
  \textbf{Important milestones in protein structure
  determination and design}
  (a) Schematic of timeline highlighting key developments
  in experimental structure determination, protein folding,
  molecular simulation, structural data resources, and
  computational protein design.
}
\label{fig2}
\end{figure*}

Deep learning, a rapidly growing subfield of machine learning, has been widely adopted in biological research in recent years due to its ability to handle heterogeneous, high-dimensional, and often noisy biological data across diverse downstream tasks \cite{ching2018opportunities,litjens2017survey,lin2023evolutionary,zhou2015predicting,wagle2026interpretable,wagle2026deep}. In protein research, its applications extend from characterizing known sequences and structures to generating new proteins for specified design objectives: (i) structure prediction methods such as AlphaFold2 and RoseTTAFold have substantially improved structural modeling, with AlphaFold2 achieving near-experimental accuracy for many CASP14 targets \cite{jumper2021highly,baek2021accurate}; (ii) protein language models (PLMs) trained on millions of sequences learn patterns relevant to protein structure and function, supporting zero-shot prediction of variant effects and sequence generation \cite{rives2021biological,meier2021language,nijkamp2023progen2}; (iii) geometry-aware generative models based on diffusion and flow matching can generate protein backbones, with methods such as RFdiffusion supporting the scaffolding of functional motifs \cite{watson2023novo,yim2023fast}; and (iv) recent models incorporate a broader range of molecular information, with AlphaFold3 predicting structures of complexes containing proteins, nucleic acids, and other molecular components, and ESM3 modeling protein sequence, structure, and functional annotations in a single framework \cite{abramson2024accurate,hayes2025simulating}. Figure~\ref{fig2} summarizes key milestones in protein structure determination and design.\\

In this work, we review generative protein design to examine how developments in modeling have expanded the range of design problems that can be addressed and where the current capabilities remain limited. We first introduce the principles of generative modeling and the representations used to encode information about the protein. Next, we review recent methods categorized by protein design task, particularly comparing how different approaches generate sequences and structures and incorporate constraints relevant to the specified function. We then discuss computational and experimental evaluation and distinguish what these assessments establish about the structure and biological activity of designed proteins. Finally, we highlight current challenges and discuss potential future directions for improving the reliability of functional protein design.

\section*{Approaches to generative modeling}\hfill

Traditionally, machine learning has been developed primarily for discriminative tasks, such as predicting a class label or regression. In contrast, generative modeling aims to approximate a data distribution so that samples can be generated from the learned model. In this section, we describe the five most commonly used approaches underpinning modern computational protein design, including variational autoencoders (VAEs), generative adversarial networks (GANs),  diffusion models, flow matching, and autoregressive transformers.

\subsection*{Variational autoencoders}\hfill
\vspace{0.3\baselineskip} 

In a standard autoencoder, the input is mapped to a deterministic latent vector, and the decoder reconstructs the output in a predictable manner. A variational autoencoder (VAE) learns a continuous, probabilistic latent space based on its training data. The encoder maps each input (e.g., a protein sequence or structure) to a distribution over latent vectors, from which samples are drawn and passed to the decoder for reconstruction. New samples can be generated by passing latent vectors sampled from the prior to the decoder (Fig. \ref{fig3}a) \cite{kingma2014auto}. \\

Specifically, the encoder $q_\phi$ maps input $x$ to a diagonal Gaussian posterior over the latent variable $z$, parameterized by mean $\mu_\phi(x)$ and variance $\sigma_\phi^2(x)$:

\[
q_\phi(z \mid x)
=
\mathcal{N}\!\left(
z;\mu_\phi(x),
\operatorname{diag}\!\left(\sigma_\phi^2(x)\right)
\right).
\]

A latent vector $z$ is then drawn via the reparameterization trick,
$z = \mu_\phi(x) + \sigma_\phi(x) \odot \epsilon$ with $\epsilon \sim \mathcal{N}(0, I)$,
which keeps sampling differentiable. The decoder $p_\theta$ then reconstructs
$x$ from $z$. Training minimizes the following weighted VAE objective:

\begin{align*}
\mathcal{L}_{\mathrm{VAE}} &= -\mathbb{E}_{q_\phi(z \mid x)}\!\left[\log p_\theta(x \mid z)\right] \\
&\quad + \beta\, D_{\mathrm{KL}}\!\left(q_\phi(z \mid x)\,\|\,p(z)\right)
\end{align*}

where the first term is the negative reconstruction log-likelihood. The second term is a
Kullback-Leibler (KL) penalty, which regularizes the posterior toward the prior
$p(z) = \mathcal{N}(0, I)$, and $\beta$ controls regularization strength. When $\beta=1$, this objective is the negative evidence
lower bound (ELBO).

\subsection*{Generative adversarial networks}\hfill
\vspace{0.3\baselineskip} 

\begin{figure*}[!htbp]
    \centering
    \includegraphics[width=\dimexpr\textwidth-7em\relax]{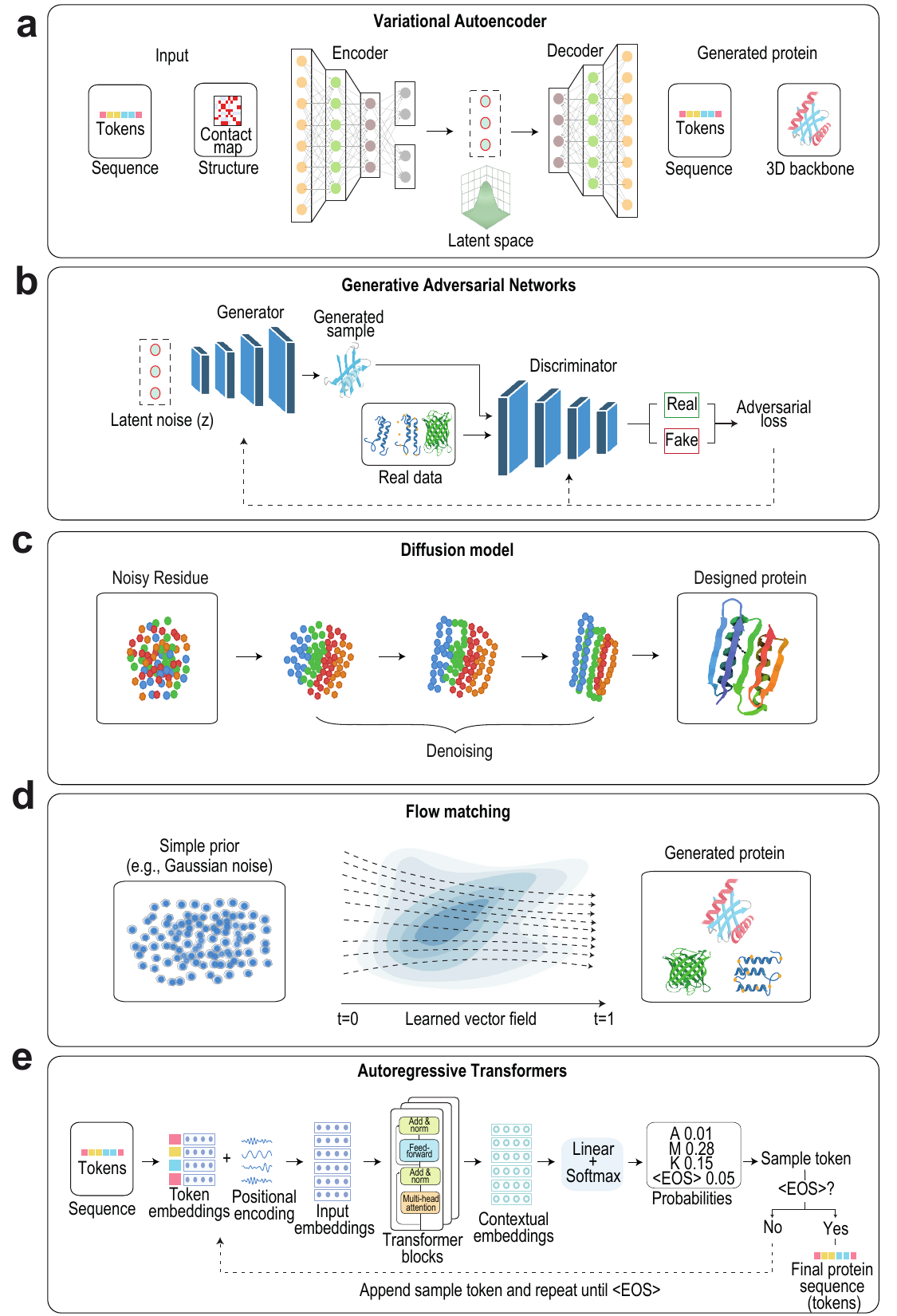}
    \caption{\textbf{Overview of common generative modeling approaches used in protein design}
    (a) Variational autoencoder - an encoder maps protein representations to a latent distribution, and a decoder reconstructs or generates protein representations from sampled latent variables. Sequence tokens and structural representations illustrate possible inputs and outputs.
    (b) Generative adversarial network - a generator produces candidates from latent noise, and a discriminator learns to distinguish generated samples from training data. Adversarial training provides feedback to improve the generator.
    (c) Diffusion model illustrating iterative denoising from a noisy representation to a generated protein structure.
    (d) Flow matching showing how a learned time-dependent vector field transforms samples from a simple prior distribution into protein representations.
    (e) Autoregressive transformer illustrating token, positional embeddings, transformer blocks, and prediction of the next-token probability distribution. Sampled tokens are appended to the sequence, and generation continues until an end-of-sequence (EOS) token is reached.}\label{fig3}
\end{figure*}

A generative adversarial network (GAN) consists of two networks (the generator and the discriminator) trained simultaneously in a competitive manner. The generator generates synthetic samples from random noise, while the discriminator is trained to distinguish these samples from real data. The generator is rewarded for fooling the discriminator, and the discriminator is rewarded when it distinguishes between real and fake data (Fig. \ref{fig3}b) \cite{NIPS2014_f033ed80}.\\

Unlike likelihood-based models, a generative adversarial network does not evaluate an explicit density. Instead, it learns the data distribution implicitly through an adversarial signal. Specifically, the generator $G$ maps a latent noise vector $z \sim p_z$ to a sample $G(z)$. The discriminator $D$ provides the probability that a given input is real. Both real samples $x \sim p_{\text{data}}$ and generated samples $G(z)$ are shown to $D$, and the two networks optimize the minimax objective:
\[
\min_{G}\max_{D}\; \mathbb{E}_{x}\!\left[\log D(x)\right] + \mathbb{E}_{z}\!\left[\log\bigl(1 - D(G(z))\bigr)\right]
\]

If $D$ becomes too confident, $G$ may receive weak gradients. Generative adversarial networks may also suffer from mode collapse, where $G$ produces a narrow range of outputs. For sequence generation (e.g., protein sequences), hard sampling of discrete tokens prevents direct backpropagation through the sampled outputs. Wasserstein-based formulations provide an alternative approach to adversarial training. For example, Gulrajani et al. proposed adding a gradient penalty to the Wasserstein GAN objective to improve training stability \cite{gulrajani2017improved}.

\subsection*{Diffusion models}\hfill
\vspace{0.3\baselineskip} 

Diffusion models learn to generate new data by reversing a process that gradually adds noise to the input (Fig. \ref{fig3}c) \cite{ho2020denoising}. Specifically, in denoising diffusion probabilistic models (DDPMs), the training input $\mathbf{x}_0$ is corrupted with Gaussian noise over $T$ steps based on a predefined noise schedule. During training, a timestep $t$ is sampled uniformly from $\{1,\ldots,T\}$, and a noisy $\mathbf{x}_t$ is constructed at the corresponding noise level. A neural network $\epsilon_\theta(\mathbf{x}_t,t)$ receives this noisy input and the timestep, and predicts the noise used to corrupt the input. Training minimizes the following noise prediction objective:

\[
\mathcal{L}_{\mathrm{DDPM}} = \mathbb{E}_{\mathbf{x}_0,\epsilon,t}\!\left[\left\|\epsilon-\epsilon_\theta(\mathbf{x}_t,t)\right\|_2^2\right],
\]

where $\epsilon \sim \mathcal{N}(\mathbf{0},\mathbf{I})$. During generation, sampling begins from $\mathbf{x}_T \sim \mathcal{N}(\mathbf{0},\mathbf{I})$. At each step, the network predicts the noise in the current state, and this prediction is used to sample the subsequent, less noisy state. Repeating this process produces the generated sample $\mathbf{x}_0$.

\subsection*{Flow matching}\hfill
\vspace{0.3\baselineskip}

A practical limitation of diffusion models is the computational cost of generating data, particularly when sampling requires many sequential denoising steps. Flow matching learns a time-dependent velocity field that moves samples from a simple prior distribution toward the data distribution (Fig. \ref{fig3}d) \cite{lipman2023flow}. At a given position and time, this field specifies the direction and speed of movement. A common training setup includes an independently drawn noise sample $\mathbf{z} \sim \mathcal{N}(\mathbf{0}, \mathbf{I})$ and a training input $\mathbf{x}$. Intermediate states are constructed by linear interpolation, $\mathbf{x}_t = (1-t)\mathbf{z} + t\mathbf{x}$, connecting noise at $t=0$ to data at $t=1$. Along this training path, the target velocity is constant and given by $\mathbf{x}-\mathbf{z}$. A neural network $v_\theta(\mathbf{x}_t,t)$ predicts this velocity from the intermediate state and its time. Training minimizes the following velocity prediction objective:

\[
\mathcal{L}_{\mathrm{FM}} = \mathbb{E}_{\mathbf{x},\mathbf{z},t}\!\left[\left\|v_\theta(\mathbf{x}_t,t)-(\mathbf{x}-\mathbf{z})\right\|_2^2\right],
\]

where the expectation is over training inputs, noise samples, and times sampled uniformly from $[0,1]$. During generation, a noise sample evolves from $t=0$ to $t=1$ by solving an ordinary differential equation (ODE) defined by the learned velocity field. The state at $t=1$ indicates the generated sample.

\subsection*{Autoregressive transformers}\hfill
\vspace{0.3\baselineskip}

Autoregressive transformers generate sequences by predicting each token from the preceding tokens. The model learns dependencies across the available sequence context by an attention mechanism. Causal masking prevents predictions from using subsequent tokens (Fig. \ref{fig3}e) \cite{vaswani2017attention}. Specifically, an input sequence (e.g., amino acid sequence) is divided into tokens, and each token is mapped to a learned embedding. Positional information is incorporated into the model to represent sequence order. These embeddings are processed through a stack of transformer blocks consisting of multi-head self-attention and feed-forward layers, with residual connections and normalization. For a token sequence $\mathbf{x} = (x_1,\ldots,x_L)$, the probability is represented as a product of conditional probabilities:

\[
p_\theta(\mathbf{x}) = \prod_{i=1}^{L} p_\theta(x_i \mid x_{<i}),
\]

where $L$ is the number of tokens and $x_{<i} = (x_1,\ldots,x_{i-1})$ denotes the preceding tokens. Training minimizes a next-token cross-entropy loss \cite{radford2018improving}. During data generation, the initial context is provided by a start token or a sequence prefix. The model outputs a probability distribution over the token vocabulary, from which the next token is sampled and appended to the sequence. This process repeats until an end-of-sequence (EOS) token is generated or a specified length is reached.

\section*{Representations for protein design}

The ability of a generative model to reason and learn from the sequence, structure, and functional annotations of proteins is determined by how this information is represented. In this section, we discuss various representation families that underlie modern generative protein models and their roles in protein design. 

\subsection*{Sequence representations}\hfill
\vspace{0.3\baselineskip}


The identity and order of amino acids contain information about the structure and function of a protein. Protein language models (PLMs) represent sequences as ordered amino acid tokens that are mapped to learned embeddings. These embeddings are then processed to obtain contextual representations, in which each residue is represented by a vector that depends on the sequence context available to the model \cite{rives2021biological}. These representations can be retained at the residue level or pooled into a single vector (e.g., by averaging) for protein-level analysis. Some models construct pairwise representations that describe relationships between residue positions and are used in structural modeling \cite{lin2023evolutionary}. For protein design, these contextual representations can be used to predict amino acid identities when generating new sequences \cite{nijkamp2023progen2}.\\

The representations also differ in the evolutionary context provided to the model. The single sequence protein language models derive their representations from the query sequence without the need for an explicit multiple sequence alignment (MSA) and rely on patterns that have been learned during pretraining. Alternatively, a multiple sequence alignment arranges evolutionarily related sequences at corresponding residue positions, with gaps indicating insertions and deletions. This provides explicit information about residue conservation and covariation between residue positions, which can be incorporated into learned representations. The information provided by a multiple sequence alignment depends on both the number and diversity of the sequences it contains, since closely related sequences can contribute redundant information \cite{rao2021msa}.

\subsection*{Structural representations}\hfill
\vspace{0.3\baselineskip}

The function of a protein is also dependent on the spatial arrangement of the residues (including contacts between residues that are distant in sequence), and such geometry can be represented at different levels. Cartesian representations specify atomic positions, from a trace of the $C_\alpha$ coordinates to representations that include side-chain atoms. A $C_\alpha$ trace captures the overall backbone shape, and the side-chain coordinates provide additional information about local arrangement and molecular contacts. Local reference frames provide another way to represent backbone geometry. Each residue is assigned a position and an orientation, commonly defined using its backbone $N$, $C_\alpha$, and $C$ atoms. These frames can be combined with torsion angles to describe side-chain conformations \cite{jumper2021highly}. Geometric information can also be represented as graphs, with atoms or residues as nodes and edges describing covalent connectivity or spatial proximity. These proximity graphs make spatial relationships accessible to a model even when the residues involved are distant in sequence \cite{jing2021learning}.\\

Alternatively, internal coordinates describe chain geometry through bond lengths and angles, including torsion angles that specify rotation around bonds. Some approaches fix bond lengths and represent the backbone using the remaining angular degrees of freedom \cite{wu2024protein}. Pairwise representations describe geometry through distances and relative orientations between residues without specifying absolute coordinates \cite{du2021trrosetta}. These quantities remain unchanged when the entire protein is rotated or translated and are therefore invariant to these global motions. Coordinates and residue frames change with the reference frame. Equivariant models process geometric quantities so that their outputs transform consistently with the input \cite{jing2021learning}. The distinction concerns how a representation behaves under global motion - invariant features can still retain information about the relative orientation of residues. For protein design, structural representations can describe the geometry being generated or provide the structural context used to select amino acid identities.

\subsection*{Multimodal representations}\hfill
\vspace{0.3\baselineskip}

Sequences and structures describe the molecular composition and geometry of a protein, while functional annotations provide information about the relevant biological activity. These sources of information can be represented jointly so that the features associated with a residue reflect both its amino acid identity and the structural and functional context. In the token-based approach, we can encode the local structural environment as a set of discrete tokens associated with each residue, and the functional annotations can be represented separately. Embeddings from the different modalities can then be combined at corresponding residue positions and processed together to obtain contextual representations \cite{hayes2025simulating}. Alternatively, modalities can be encoded separately and aligned within a shared embedding space through contrastive learning. This encourages representations of matching observations, such as a protein sequence and its functional description, to be more similar than those of mismatched pairs. The resulting alignment enables comparison of information expressed across different modalities (for example, when retrieving proteins using a textual description of a biological activity) \cite{su2025trimodal}. For protein design, a text-derived representation can be mapped to a protein representation and used to condition sequence generation. Such design could allow functional descriptions to guide the generation of candidate protein sequences \cite{liu2025text}.

\section*{Designing proteins with generative models}


A key goal of protein design is to identify amino acid sequences and three-dimensional structures that can carry out a specified biological function. Sequences can be generated without a predefined backbone, or a backbone can be generated first and used to guide sequence design. Alternatively, both sequence and structure can be designed together. More constrained objectives include constructing a scaffold around a specified motif or designing a protein to interact with a target. In this section, we summarize key generative approaches by design task, including sequence generation, backbone and structural generation, structure-conditioned sequence design, joint sequence-structure design, motif scaffolding, and binder and antibody design. Figures~\ref{fig:sequence-backbone-design} and~\ref{fig:joint-constrained-design} summarize a total of 59 distinct generative methods across these tasks and compare them by generation approach, modality, architecture, and conditioning.

\begin{figure*}[!htbp]
    \centering
    \includegraphics[width=\textwidth-0.9em\relax]{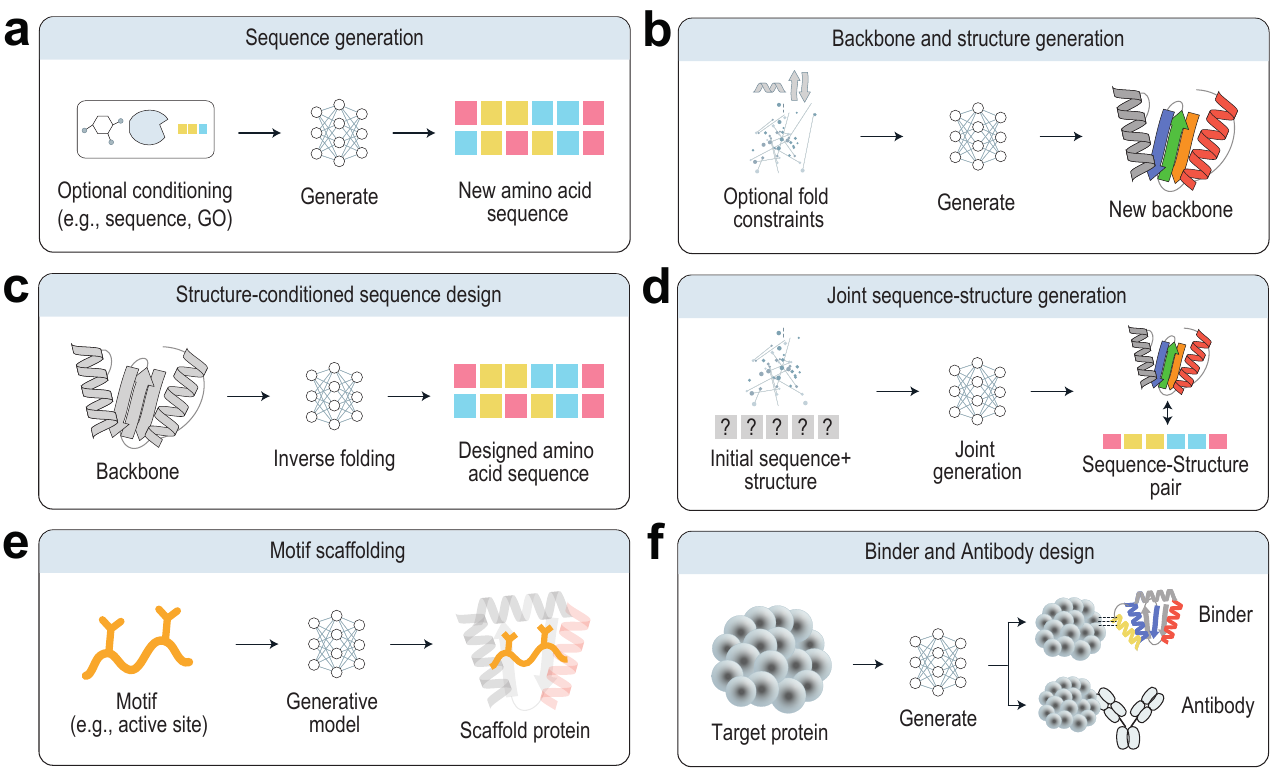}
    \caption{\textbf{Overview of generative tasks in protein design}
    (a) Generation of new amino acid sequences with optional conditioning inputs (e.g., existing sequences, Gene Ontology annotations, natural language descriptions).
    (b) Generation of new protein backbones with optional conditioning inputs (e.g., fold specifications, symmetry constraints, input sequences).
    (c) Inverse folding of a specified backbone to identify a compatible amino acid sequence.
    (d) Joint generation of an amino acid sequence and its corresponding three-dimensional structure.
    (e) Generation of a protein scaffold around a specified motif (e.g., an active site), while preserving the geometry of the motif.
    (f) Design of a protein binder or antibody against a specified target protein.}
    \label{fig4tasks}
\end{figure*}

\subsection*{Sequence generation}\hfill
\vspace{0.3\baselineskip}


Generative models can produce new amino acid sequences by learning from natural sequence variation, without requiring a predefined backbone (Fig. \ref{fig4tasks}a). Hawkins-Hooker et al. compared an aligned-sequence variational autoencoder with a hybrid autoregressive variant trained on unaligned sequences. The authors then generated luxA variants around the latent encoding of a reference sequence, and some variants from each model were found to retain measurable luminescence \cite{hawkins2021generating}. ProteinGAN learned variation within the malate dehydrogenase family through adversarial training. Of the 55 generated sequences tested experimentally, 13 produced soluble proteins with measurable catalytic activity \cite{repecka2021expanding}. PoET adopts a different approach to family-directed generation by using related protein sequences as conditioning in a model trained across families. It can therefore generate additional members of a specified family without separate family-specific fine-tuning \cite{truong2023poet}.\\

\begin{figure*}[p]
\centering
\fontsize{8.25}{9.1}\selectfont
\setlength{\tabcolsep}{1.85pt}
\setlength{\arrayrulewidth}{.45pt}
\setlength{\extrarowheight}{0pt}
\renewcommand{\arraystretch}{1.16}
\begin{adjustbox}{max totalsize={\textwidth}{.88\textheight}}
\begin{tabular}{|
D{1.68cm} E{2.10cm} D{.80cm} E{2.72cm}
D{.45cm} D{.38cm} D{.38cm} D{.42cm}
E{2.82cm} E{3.34cm} D{.73cm}@{}|}
\PDheader
\PDgroup{14}{Sequence\\generation} \PDrow{Hawkins-Hooker et al.}{hawkins2021generating}{Variational generation}{s}{MSA VAE / autoregressive VAE}{Reference LuxA (latent seed)}{\PDgh{alex-hh/deep-protein-generation}}
\PDrow{ProteinGAN}{repecka2021expanding}{Adversarial generation}{s}{Self-attention GAN}{Unconditional}{\PDgh{Biomatter-Designs/ProteinGAN}}
\PDrow{PoET}{truong2023poet}{Autoregressive decoding}{s}{Protein-family transformer}{Optional related sequences}{\PDgh{OpenProteinAI/PoET}}
\PDrow{ProtGPT2}{ferruzprotgpt2}{Autoregressive decoding}{s}{GPT-2 transformer}{Unconditional}{\PDcode{https://huggingface.co/nferruz/ProtGPT2}}
\PDrow{ProGen2}{nijkamp2023progen2}{Autoregressive decoding}{s}{Transformer}{Optional prefix}{\PDgh{salesforce/progen/blob/main/progen2/README.md}}
\PDrow{ProGen3}{bhatnagar2026scaling}{Autoregressive decoding}{s}{MoE transformer}{Optional flanking sequence}{\PDgh{Profluent-AI/progen3}}
\PDrow{EvoDiff}{alamdari2023protein}{Discrete diffusion}{s}{Dilated CNN (sequence model)}{Optional sequence context}{\PDgh{microsoft/evodiff}}
\PDrow{DPLM}{wang2024diffusion}{Discrete diffusion}{s}{ESM-2--style transformer}{Optional sequence context}{\PDgh{bytedance/dplm}}
\PDrow{DiMA}{meshchaninov2024diffusion}{Latent diffusion}{s}{PLM encoder + transformer denoiser}{Unconditional}{\PDgh{MeshchaninovViacheslav/DiMA}}
\PDrow{ProtBFN}{atkinson2025protein}{Bayesian flow networks}{s}{Transformer}{Unconditional}{\PDgh{instadeepai/protein-sequence-bfn}}
\PDrow{ProGen}{madani2023large}{Autoregressive decoding}{sf}{Transformer}{Property tags}{\PDgh{salesforce/progen}}
\PDrow{ZymCTRL}{munsamy2022zymctrl}{Autoregressive decoding}{sf}{CTRL transformer}{EC number}{\PDcode{https://huggingface.co/AI4PD/ZymCTRL}}
\PDrow{ProteoGAN}{kucera2022conditional}{Conditional Wasserstein GAN}{sf}{Convolutional generator and critic}{GO labels}{\PDgh{timkucera/proteogan}}
\PDrow{ProteinDT}{liu2025text}{Autoregressive or diffusion decoding}{st}{Text/protein encoders + decoder}{Natural language}{\PDgh{chao1224/ProteinDT}}
\PDseparator
\PDgroup{13}{Backbone\\and\\structure\\generation} \PDrow{ProteinSGM}{lee2023score}{Score-based diffusion}{r}{Score network on geometry maps}{Unconditional generation}{\PDcode{https://gitlab.com/mjslee0921/proteinsgm}}
\PDrow{FoldingDiff}{wu2024protein}{Angle diffusion}{r}{Transformer}{Specified backbone length}{\PDgh{microsoft/foldingdiff}}
\PDrow{ProtDiff}{trippe2022diffusion}{Coordinate diffusion}{r}{EGNN (C$\alpha$)}{Unconditional}{\PDgh{blt2114/ProtDiff_SMCDiff}}
\PDrow{Genie}{lin2023generating}{Coordinate diffusion}{r}{SE(3) denoiser}{Unconditional}{\PDgh{aqlaboratory/genie}}
\PDrow{FrameDiff}{yim2023se}{Frame diffusion}{r}{SE(3) equivariant net}{Unconditional}{\PDgh{jasonkyuyim/se3_diffusion}}
\PDrow{RFdiffusion}{watson2023novo}{Frame diffusion}{r}{RoseTTAFold}{Optional fold or symmetry}{\PDgh{RosettaCommons/RFdiffusion}}
\PDrow{Chroma}{ingraham2023illuminating}{Correlated diffusion}{st}{SE(3) graph networks}{Optional symmetry, shape or text}{\PDgh{generatebio/chroma}}
\PDrow{RFdiffusion3}{butcher2026atomic}{Free atom diffusion}{sr}{Atom-level transformer U-Net}{Optional atomic constraints}{\PDgh{RosettaCommons/foundry}}
\PDrow{FrameFlow}{yim2023fast}{SE(3) flow matching}{r}{SE(3) equivariant net}{Unconditional}{\PDgh{microsoft/protein-frame-flow}}
\PDrow{FoldFlow}{bose2024se}{SE(3) flow matching}{r}{SE(3) equivariant net}{Unconditional}{\PDgh{DreamFold/FoldFlow}}
\PDrow{FoldFlow-2}{huguet2024sequence}{SE(3) flow matching}{sr}{SE(3) equivariant network + ESM-2}{Optional sequence}{\PDgh{DreamFold/FoldFlow}}
\PDrow{Genie 2}{lin2024out}{Coordinate diffusion}{r}{SE(3) equivariant net}{No motif at sampling}{\PDgh{aqlaboratory/genie2}}
\PDrow{Proteina}{geffner2025proteina}{Coordinate flow matching}{r}{Scalable transformer}{Optional CATH fold labels}{\PDgh{NVIDIA-BioNeMo/proteina}}
\PDseparator
\PDgroup{10}{Structure-\\conditioned\\sequence\\design} \PDrow{ProteinMPNN}{dauparas2022robust}{Autoregressive decoding}{sr}{MPNN}{Fixed backbone}{\PDgh{dauparas/ProteinMPNN}}
\PDrow{ESM-IF1}{hsu2022learning}{Autoregressive decoding}{sr}{GVP-Transformer}{Fixed backbone}{\PDgh{facebookresearch/esm}}
\PDrow{CarbonDesign}{ren2024accurate}{MRF sequence decoding}{sr}{Inverseformer + ESM-2 + MRF}{Fixed backbone}{\PDgh{zhanghaicang/carbonmatrix_public}}
\PDrow{LM-Design}{zheng2023structure}{Iterative masked decoding}{sr}{PLM + structural adapter}{Fixed backbone}{\PDgh{BytedProtein/ByProt}}
\PDrow{DPLM with structural adapter}{wang2024diffusion}{Discrete diffusion}{sr}{DPLM + structural adapter}{Fixed backbone}{\PDgh{bytedance/dplm}}
\PDrow{gcWGAN}{karimi2020novo}{Conditional WGAN}{sr}{Generator + fold predictor}{Target fold embedding}{\PDgh{Shen-Lab/gcWGAN}}
\PDrow{AF2-design}{goverde2023novo}{Gradient or MCMC optimization}{sr}{Frozen AlphaFold2}{Target backbone}{\PDgh{bene837/af2seq}}
\PDrow{LigandMPNN}{dauparas2025atomic}{Autoregressive decoding}{sr}{MPNN + atomic context encoder}{Backbone and nearby atoms}{\PDgh{dauparas/LigandMPNN}}
\PDrow{ADFLIP}{yi2025all}{Discrete flow matching}{sr}{Atom/residue GNN + attention decoder}{Backbone and atomic context}{\PDgh{ykiiiiii/ADFLIP}}
\PDrow{DynamicMPNN}{abrudan2026multi}{Autoregressive decoding}{sr}{GVP-GNN}{Multiple conformations}{\PDgh{Alex-Abrudan/DynamicMPNN}}
\hline
\end{tabular}
\end{adjustbox}
\caption{\textbf{Generative methods for protein sequence and structure design.} Methods are grouped by sequence generation, backbone and structure generation, and structure-conditioned sequence design. Colored dots indicate the modalities used or generated for the task shown. Seq: amino acid sequence, Str: structure, Fn: structured functional annotation, Txt: natural-language text. Implementation icons link to the method repository.}
\label{fig:sequence-backbone-design}
\end{figure*}

Methods, including ProtGPT2 and ProGen2, are trained on sequences from diverse protein families and can generate sequences without a supplied backbone. The authors further assessed candidates generated through computational analyses of sequence and structural properties \cite{ferruzprotgpt2,nijkamp2023progen2}. ProGen3 additionally supports infilling internal sequence regions using surrounding residues as context, and selected proteins were evaluated experimentally using a split-GFP expression assay \cite{bhatnagar2026scaling}. The same general generation task can also be approached through denoising. EvoDiff and DPLM learn to recover amino acid identities from corrupted sequences and support generation without a target structure and completion around retained sequence regions \cite{alamdari2023protein,wang2024diffusion}. DiMA performs diffusion in continuous representations produced by pretrained protein encoders and decodes the generated representations into amino acid sequences \cite{meshchaninov2024diffusion}. ProtBFN generates sequences by iteratively updating distributions over token identities through Bayesian inference and neural network predictions \cite{atkinson2025protein}.\\

Functional labels and text descriptions can also be used to guide sequence generation. ProGen incorporates control tags describing protein properties and can be fine-tuned using sequences from a family of interest. Fine-tuning on five lysozyme families generated proteins with experimentally demonstrated catalytic activity \cite{madani2023large}. ZymCTRL uses Enzyme Commission numbers to specify the requested enzyme class \cite{munsamy2022zymctrl}, while ProteoGAN conditions generation on combinations of molecular function labels from the Gene Ontology \cite{kucera2022conditional}. ProteinDT allows desired properties to be expressed through natural-language descriptions, which serve as conditioning information for autoregressive or diffusion-based sequence decoders. Additionally, it supports text-guided editing of existing sequences through optimization or interpolation in the learned representation space \cite{liu2025text}.

\subsection*{Backbone and structure generation}\hfill
\vspace{0.3\baselineskip}


Designing the protein backbone first offers a way to explore possible folds before selecting an amino acid sequence (Fig. \ref{fig4tasks}b). ProteinSGM achieves this by generating inter-residue distance and orientation maps through score-based diffusion, followed by constrained Rosetta minimization to obtain three-dimensional backbones \cite{lee2023score}. FoldingDiff instead generates backbone bond and dihedral angles through denoising and reconstructs the corresponding atomic coordinates using fixed bond lengths \cite{wu2024protein}. ProtDiff and Genie operate directly on $C_\alpha$ coordinates, with Genie additionally deriving local reference frames from the noisy coordinates to guide denoising \cite{trippe2022diffusion,lin2023generating}. FrameDiff models both residue positions and orientations through a score-based diffusion process over rigid residue frames and learns to generate backbones without relying on a pretrained structure prediction network \cite{yim2023se}.\\

Structural requirements can also guide the generation of candidate backbones. RFdiffusion adapts the pretrained RoseTTAFold network to denoise protein structures and supports both generation without a supplied backbone and generation conditioned on secondary-structure or fold specifications. Compatible amino acid sequences are subsequently designed for the generated backbones using ProteinMPNN \cite{watson2023novo,dauparas2022robust}. Chroma uses a correlated noise process informed by the spatial statistics of protein chains and supports geometric requirements, including symmetry and overall shape, to be incorporated during sampling. A separate design network generates sequences and side-chain conformations conditioned on the sampled backbone. Crystal structures of two designed proteins closely matched the generated models \cite{ingraham2023illuminating}. RFdiffusion3 further supports structural design under constraints on individual atoms through free-atom diffusion \cite{butcher2026atomic}.\\

Sampling efficiency can affect how many candidate backbones are explored. Methods such as FrameFlow and FoldFlow learn flows over residue positions and orientations to generate protein backbones. FrameFlow demonstrated generation with fewer sampling steps than FrameDiff in the reported comparison. In contrast, FoldFlow introduced deterministic and stochastic formulations, including variants that use optimal-transport couplings \cite{yim2023fast,bose2024se}. FoldFlow-2 incorporates sequence representations from a protein language model and is trained with and without sequence information and supports both unconditional and sequence-conditioned backbone generation \cite{huguet2024sequence}. Another approach to extending structural coverage is to train on larger collections of predicted structures. Genie 2 uses a confidence- and length-filtered subset of structural cluster representatives from the AlphaFold Protein Structure Database to broaden its training coverage \cite{lin2024out}. Proteina combines large-scale training with flow matching over $C_\alpha$ coordinates and hierarchical fold-class labels that guide the secondary-structure composition and topology of generated backbones. The authors reported generating backbones up to 800 residues long, which were assessed computationally using sequence design and structure prediction \cite{geffner2025proteina}.

\subsection*{Structure-conditioned sequence design}\hfill
\vspace{0.3\baselineskip}


Inverse folding aims to identify amino acid sequences that can adopt a given protein backbone, and several distinct sequences may be compatible with the same backbone (Fig. \ref{fig4tasks}c). ProteinMPNN represents the backbone as a graph of spatially neighboring residues and samples amino acid identities autoregressively. In one application, a redesigned interface in a tetrahedral assembly closely matched the intended geometry in the crystal structure \cite{dauparas2022robust}. ESM-IF1 also decodes sequences autoregressively, combining a geometric encoder with a transformer decoder and supplementing experimentally determined training structures with AlphaFold2 predictions for millions of sequences \cite{hsu2022learning}. CarbonDesign models both individual amino acid identities and their pairwise dependencies using a Markov random field, whose parameters are inferred from the backbone \cite{ren2024accurate}. Pretrained protein language models can also be adapted to perform inverse folding. LM-Design and the structure-conditioned version of DPLM use structural adapters to incorporate backbone information into sequence decoding \cite{zheng2023structure,wang2024diffusion}.\\

A structural target can guide sequence design either by representing the desired fold or by using a loss that measures agreement with the target geometry. gcWGAN conditions sequence generation on a low-dimensional representation derived from similarities to reference folds and uses feedback from a sequence-to-fold predictor during training \cite{karimi2020novo}. In AF2-design, AlphaFold2 predicts structures for candidate sequences, and differences from the target structure guide sequence updates through gradient descent or Monte Carlo search. Following a separate Rosetta surface redesign step, selected designs exhibited folding in circular dichroism experiments \cite{goverde2023novo}. At a binding site, the choice of amino acid depends on the geometry of the backbone and the atoms that the side chain will contact. LigandMPNN incorporates nearby non-protein atoms, including small molecules, nucleic acids, and metal ions, into the sequence design step. A subsequent packing model predicts side-chain conformations for the designed sequence \cite{dauparas2025atomic}. ADFLIP incorporates predicted side chains into the structural context as residues are assigned during sequence generation by discrete flow matching \cite{yi2025all}. For designs that accommodate multiple conformations, DynamicMPNN encodes the provided structures and pools features at corresponding residue positions before decoding a sequence compatible with those states \cite{abrudan2026multi}.

\subsection*{Joint sequence-structure design}\hfill
\vspace{0.3\baselineskip}


In a joint design, both the amino acid identities and three-dimensional structure of the protein can be modeled during generation (Fig. \ref{fig4tasks}d). ProteinGenerator applies Gaussian noise to a numerical representation of the sequence and adapts RoseTTAFold to predict the denoised sequence and its structure at each step. The crystal structure of a designed five-repeat protein closely matched the design \cite{lisanza2025multistate}. CarbonNovo uses information between a backbone diffusion model and a sequence model using Markov Random Fields at each sampling step, while Multiflow couples continuous flows for backbone geometry with a discrete flow for amino acid identities \cite{ren2024carbonnovo,campbell2024generative}. Generating side chains as well as the backbone introduces difficulties because the number and arrangement of side-chain atoms depend on the residue identity. Protpardelle retains candidate coordinates for alternative side chains and uses sequence predictions during denoising to select the coordinates to update \cite{chu2024all}. La-Proteina jointly generates $\mathrm{C}_{\alpha}$ coordinates and per-residue latent vectors. A decoder then uses both to recover the amino acid sequence and the coordinates of the remaining heavy atoms. Proteina Atomistica models the backbone, amino acid identities, and side-chain geometry directly \cite{geffner2026proteina,reidenbach2025consistent}.\\

\begin{figure*}[p]
\centering
\fontsize{8.25}{9.1}\selectfont
\setlength{\tabcolsep}{1.85pt}
\setlength{\arrayrulewidth}{.45pt}
\setlength{\extrarowheight}{1.4pt}
\renewcommand{\arraystretch}{1.16}
\begin{adjustbox}{max totalsize={\textwidth}{.88\textheight}}
\begin{tabular}{|
D{1.68cm} E{2.10cm} D{.80cm} E{2.72cm}
D{.45cm} D{.38cm} D{.38cm} D{.42cm}
E{2.82cm} E{3.34cm} D{.73cm}@{}|}
\PDheader
\PDgroup{10}{Joint\\sequence--\\structure\\design} \PDrow{ProteinGenerator}{lisanza2025multistate}{Sequence diffusion}{sr}{Adapted RoseTTAFold}{Optional structural constraints}{\PDgh{RosettaCommons/protein_generator}}
\PDrow{CarbonNovo}{ren2024carbonnovo}{Coupled diffusion and sequence sampling}{sr}{Backbone denoiser + sequence model}{Unconditional}{\PDgh{CarbonMatrixLab/carbonnovo}}
\PDrow{Multiflow}{campbell2024generative}{Discrete--continuous flow matching}{sr}{SE(3) equivariant net}{Unconditional}{\PDgh{jasonkyuyim/multiflow}}
\PDrow{Protpardelle}{chu2024all}{All-atom diffusion}{sr}{U-ViT + non-AR mini-MPNN}{Unconditional}{\PDgh{ProteinDesignLab/protpardelle}}
\PDrow{La-Proteina}{geffner2026proteina}{Partially latent flow matching}{sr}{Transformer + atomistic VAE decoder}{Unconditional}{\PDgh{NVIDIA-BioNeMo/la-proteina}}
\PDrow{Proteina Atomistica}{reidenbach2025consistent}{Discrete--continuous flow matching}{sr}{Residue- and atom-level transformers}{Unconditional}{\textemdash}
\PDrow{ESM3}{hayes2025simulating}{Masked token generation}{srf}{Multitrack transformer}{Sequence, structure or function prompts}{\PDgh{evolutionaryscale/esm}}
\PDrow{DPLM-2}{wang2025dplm}{Discrete diffusion}{sr}{Multimodal transformer}{Optional token context}{\PDgh{bytedance/dplm}}
\PDrow{CoFlow}{yang2025co}{Discrete flow matching}{sr}{ESM3-based transformer}{Optional token context}{\PDgh{LtECoD/CoFlow}}
\PDrow{CodeFP}{chen2026co}{Discrete diffusion}{srf}{DPLM-2--based transformer}{GO terms and retrieved motifs}{\PDgh{PharMolix/OpenBioMed/blob/main/skills/functional-protein-design/SKILL.md}}[1.5pt]
\PDseparator
\PDgroup{9}{Motif\\scaffolding} \PDrow{RFdiffusion}{watson2023novo}{Frame diffusion}{sr}{RoseTTAFold}{Motif geometry and residue identities}{\PDgh{RosettaCommons/RFdiffusion}}
\PDrow{Genie 2}{lin2024out}{Coordinate diffusion}{sr}{SE(3) equivariant net}{Motif sequence and geometry}{\PDgh{aqlaboratory/genie2}}
\PDrow{SMCDiff using ProtDiff}{trippe2022diffusion}{Diffusion + SMC sampling}{r}{ProtDiff (EGNN)}{Structural motif}{\PDgh{blt2114/ProtDiff_SMCDiff}}
\PDrow{FrameFlow with motif guidance}{yim2024improved}{Guided flow sampling}{r}{FrameFlow denoiser}{Structural motif}{\PDgh{microsoft/protein-frame-flow}}
\PDrow{PAR}{qu2026protein}{Autoregressive generation + flow decoding}{r}{Coarse-to-fine transformer}{Motif coordinates}{\PDgh{ByteDance-Seed/par-protein}}
\PDrow{Riff-Diff}{braun2026computational}{Fragment search + diffusion}{sr}{RFdiffusion denoiser}{Catalytic groups}{\PDgh{mabr3112/riff_diff_protflow}}
\PDrow{RFdiffusion2}{ahern2026atom}{Atom and frame flow matching}{sr}{RoseTTAFold All-Atom}{Catalytic atom geometry}{\PDgh{RosettaCommons/RFdiffusion2}}
\PDrow{RFdiffusion3}{butcher2026atomic}{Free atom diffusion}{sr}{Atom-level transformer U-Net}{Catalytic atoms}{\PDgh{RosettaCommons/foundry}}
\PDrow{CaGE}{qian2026complex}{Iterative guided redesign}{sr}{RoseTTAFold3 + LigandMPNN}{Initial scaffold and active-site geometry}{\PDgh{baker-laboratory/CaGE}}[1.5pt]
\PDseparator
\PDgroup{9}{Binder and\\antibody\\design} \PDrow{RFdiffusion}{watson2023novo}{Frame diffusion}{sr}{RoseTTAFold}{Target sequence and structure, optional hotspots}{\PDgh{RosettaCommons/RFdiffusion}}
\PDrow{BindCraft}{pacesa2025one}{Gradient optimization + sequence redesign}{sr}{AF2-Multimer + soluble ProteinMPNN}{Target structure}{\PDgh{martinpacesa/BindCraft}}
\PDrow{Proteina-Complexa}{didi2026scaling,didi2026latent}{Partially latent flow + search}{sr}{La-Proteina--based transformer}{Target sequence/structure, optional hotspots}{\PDgh{NVIDIA-BioNeMo/Proteina-Complexa}}
\PDrow{RefineGNN}{jin2021iterative}{Autoregressive decoding + refinement}{sr}{Graph neural network}{Antibody sequence context}{\PDgh{wengong-jin/RefineGNN}}
\PDrow{DiffAb}{luo2022antigen}{Joint diffusion}{sr}{Geometric denoiser}{Framework and antigen}{\PDgh{luost26/diffab}}
\PDrow{RFantibody}{bennett2026atomically}{Frame diffusion + sequence decoding}{sr}{Fine-tuned RFdiffusion + ProteinMPNN}{Framework and epitope}{\PDgh{RosettaCommons/RFantibody}}
\PDrow{Germinal}{mille2026efficient}{Gradient-based optimization}{sr}{AF2-Multimer + IgLM}{Framework and target}{\PDgh{SantiagoMille/germinal}}
\PDrow{BoltzGen}{stark2025boltzgen}{All-atom diffusion}{sr}{Boltz-2--style trunk and denoiser}{Target and design constraints}{\PDgh{HannesStark/boltzgen}}
\PDrow{Prompteina}{lee2026designing}{Partially latent flow matching}{srt}{Proteina-Complexa + text encoders}{Target structure and text prompt}{\PDgh{jleechung/prompteina}}
\hline
\end{tabular}
\end{adjustbox}
\caption{\textbf{Generative methods for joint and constrained protein design.} Methods are grouped by joint sequence-structure design, motif scaffolding, and binder and antibody design. Colored dots indicate the modalities used or generated for the task shown. Seq: amino acid sequence, Str: structure, Fn: structured functional annotation, Txt: natural-language text. Implementation icons link to the method repository. A dash indicates that an implementation link could not be verified.}
\label{fig:joint-constrained-design}
\end{figure*}

Representing backbone geometry as tokens allows both sequence and structure to be generated in discrete space. ESM3 predicts masked sequence and structure tokens from the available context. During the design of esmGFP, the authors alternated between sequence and structure updates and measured fluorescence experimentally \cite{hayes2025simulating}. DPLM-2 jointly denoises amino acid and backbone tokens, while CoFlow uses discrete flow matching to generate both token types  \cite{wang2025dplm,yang2025co}. CodeFP uses Gene Ontology terms to guide the generation of both sequence and structure tokens and assesses the specified functions computationally \cite{chen2026co}.

\subsection*{Motif scaffolding}\hfill
\vspace{0.3\baselineskip}


Motif scaffolding uses the geometry of a region such as an enzyme active site to guide the design of the surrounding protein (Fig. \ref{fig4tasks}e). The motif can span distant parts of the sequence, and the scaffold must hold the corresponding residues in the specified three-dimensional space. RFdiffusion takes the backbone frames of the motif as input and denoises the surrounding backbone \cite{watson2023novo}. Genie 2 can scaffold several motifs in one protein while allowing their relative positions and orientations to change at inference time \cite{lin2024out}. A motif can also be introduced at the sampling step using a model trained to generate complete backbones. SMCDiff uses particle filtering to generate scaffolds compatible with the supplied motif, while the guidance variant of FrameFlow steers sampling toward the desired motif geometry \cite{trippe2022diffusion,yim2024improved}. PAR instead aligns and inserts the supplied motif coordinates at successive resolutions of coarse-to-fine backbone generation without motif-specific fine-tuning \cite{qu2026protein}.\\

For an enzyme active site, the positions of the backbone atoms alone may leave the geometry of the catalytic side chains poorly constrained. Riff-Diff holds the catalytic functional groups in place while searching for helical fragments that can support them. It then uses RFdiffusion to build backbones around those fragments \cite{braun2026computational}. RFdiffusion2 takes the coordinates of selected catalytic atoms as input and samples the side-chain conformations and sequence positions of the corresponding residues as it generates the backbone. The authors showed that a designed cysteine hydrolase showed ester cleavage in the biochemical assay \cite{ahern2026atom}. RFdiffusion3 uses diffusion over individual atoms and was used to design phosphotriesterases with experimentally measured activity \cite{butcher2026atomic}. A related challenge is to find a sequence that folds into the intended geometry at the active site. In a recent study, CaGE alternates between guided structure prediction and sequence redesign, reduces the strength of the geometric guidance over successive rounds, and the authors confirmed retro-aldolase activity in one of the purified designs \cite{qian2026complex}.

\subsection*{Binder and antibody design}\hfill
\vspace{0.3\baselineskip}


The surface of a target protein constrains the geometry and chemistry of the binding interface, while the sequence and fold of the binder can be designed around it (Fig. \ref{fig4tasks}f). RFdiffusion generates a binder backbone conditioned on the target structure and can use selected target residues to indicate the intended binding site. ProteinMPNN then designs a sequence for the generated backbone \cite{watson2023novo}. BindCraft uses gradients from AlphaFold2 multimer to optimize a binder sequence for the predicted interface and redesigns residues outside the interface using a soluble variant of ProteinMPNN. For a binder targeting the dust mite allergen Der f7, the crystal structure of the complex closely matched the design \cite{pacesa2025one}. Proteina-Complexa extends La-Proteina to generate a binder sequence and its atomic structure jointly in the context of a target. During sampling, predicted interface scores can guide the search toward candidate binders \cite{didi2026scaling}. Another recent study used reward-guided search at inference time with Proteina-Complexa to design binders to protein targets and a carbohydrate, and experimentally confirmed their binding \cite{didi2026latent}.\\

Antibody design often starts from a chosen framework and focuses on the complementarity-determining regions that make many of the contacts with the antigen. RefineGNN uses the surrounding antibody sequence to assign CDR residues and iteratively refine their backbone coordinates \cite{jin2021iterative}. DiffAb jointly denoises CDR residue identities and orientations within a supplied antibody framework and antigen structure \cite{luo2022antigen}. RFantibody uses a fine-tuned RFdiffusion model to generate CDR backbones and sample how the antibody framework is positioned relative to a specified epitope. During experimental screening, Cryo-EM microscopy confirmed the binding poses of selected antibody fragments \cite{bennett2026atomically}. Germinal combines gradients from AlphaFold multimer with guidance from an antibody language model to optimize CDR sequences and selected designs were able to bind their targets with nanomolar affinities \cite{mille2026efficient}. BoltzGen uses the same all-atom diffusion model to design protein binders, nanobodies, and peptides, with inputs that can specify binding sites and fixed structural regions \cite{stark2025boltzgen}. Prompteina takes a target structure and uses a text prompt to guide properties such as the binder fold or the interface chemistry. The authors reported six designs with clear submicromolar affinities for GM2A measured by surface plasmon resonance \cite{lee2026designing}. Across the above-discussed tasks, evaluation therefore requires criteria appropriate to the design objective, particularly clear distinctions between in silico predictions and experimental validation of the intended structural or functional properties.

\section*{Evaluating generative protein design}

Computational evaluation of designed proteins looks at sequence-structure compatibility while maintaining the design constraints and the diversity or novelty of the candidates generated. For sequences designed without a specified backbone, protein language model scores and structure predictions are used to assess sequence plausibility and structural confidence \cite{pmlr-v306-kuang26d}. For sequence design conditioned on a structure, sequence recovery is used to measure the proportion of amino acid identities that are recovered in the designed sequence compared to a reference sequence. This does not necessarily indicate whether a different sequence could in fact fold to the same backbone \cite{dauparas2022robust}. For generated protein structures, self-consistency is assessed by predicting the structure of the designed sequence and comparing it to the intended structure using metrics such as root mean square deviation (RMSD) or the Template Modeling (TM) score. The nature of such assessment will depend on the sequence design and structure prediction methods used \cite{chu2024all}. For joint generation, the evaluation of the generated sequence-structure pair should be distinct from the evaluation of the structure following sequence redesign. Prediction confidence can provide useful complementary information, though the measures used (e.g., predicted local distance difference test) are proxies for structural accuracy rather than biological activity \cite{jumper2021highly}. An evaluation of experimentally characterized monomer designs found that AlphaFold, ESMFold, and ProteinMPNN showed only moderate ability to distinguish successful from unsuccessful designs, with overlapping confidence scores and differences across structural topologies \cite{garcia2026evaluating}.\\

Motif scaffolding requires assessment of motif preservation within the predicted scaffold. For example, MotifBench uses separate RMSD criteria to assess whether the predicted structure recovers the supplied motif and whether it agrees with the generated backbone overall \cite{zheng2025motifbench}. These comparisons distinguish recovery of the local geometry from recovery of the surrounding scaffold. Structures containing backbone and side-chain coordinates also require assessment of bond lengths, angles, steric clashes, backbone and side-chain torsions \cite{zheng2025motifbench,chu2024all,chen2010molprobity}. When catalytic side chain atoms are specified, their geometry should also be evaluated, since backbone agreement alone does not establish preservation of the catalytic arrangement. For binder design, interface confidence scores can be used to inform candidate selection, but binding affinity and specificity will still need to be assessed experimentally \cite{pacesa2025one}. When a binding site is specified, the contacts in the predicted complex can also be examined to assess whether the predicted binding pose is consistent with the intended site.\\

Comparisons between methods must be made under identical task definitions and with explicit reporting of sampling budgets, filtering criteria, evaluation models and success criteria. Sampling budgets should include the number of sequences evaluated per backbone and any repeated prediction or refinement steps. Comparisons should also distinguish improvements in the generator from those introduced by downstream sequence redesign and filtering. The evaluation of generated candidates must be able to differentiate between diversity within the sampled set and novelty relative to a reference dataset. MotifBench takes a common scaffold evaluation approach where a fixed number of sequences are evaluated per backbone and successful scaffolds are clustered, and the results are compared to Protein Data Bank structures. The authors also discuss how memorization by the sequence design and structure prediction models could inflate motif preservation results \cite{zheng2025motifbench}. Benchmark construction should therefore take into account sequence and structural similarity to the training data of both the generator and the evaluation models where such data are available. For function-conditioned generation, PDFBench evaluates the designs using descriptions or functional keywords and looks at sequence plausibility, structural properties, and agreement with the functional information provided. These computational proxies should be distinguished from experimental evaluation of the specified activity \cite{pmlr-v306-kuang26d}.\\

Experimental assessment introduces another distinction between the properties that are measured and the candidates that are actually tested. Expression, structural agreement, binding, and catalytic activity make different claims about a design, as illustrated in the separate structural and functional experiments performed in the ProteinMPNN study \cite{dauparas2022robust}. For binder design, RFdiffusion assessed binding using biolayer interferometry, while Cryo-EM of a selected influenza hemagglutinin binder complex showed close agreement with the designed interaction geometry. The authors also found that computational filtering was a factor in the experimental success rate \cite{watson2023novo}. Therefore, the hit rates reported need to state the assay criterion, the number of candidates that were tested, and the selection and optimization steps that occurred prior to the measurement. The numbers that are generated and filtered and the numbers that are actually tested experimentally should be recorded in order to enable assessment of both the yield of the design process and the performance of the candidates that were selected. Further, assay conditions, controls, replicate measurements, and uncertainty in reported hit rates should also be provided where appropriate. Hit rates among tested candidates remain conditional on the selection process, and candidates that were not tested cannot be classified as experimental failures.

\section*{Challenges and future directions}

An important issue with generative protein design is learning a sufficiently wide range of sequence-structure and function relationships in order to be able to design proteins outside of the well-represented examples. Simply increasing the number of generated candidates does not necessarily result in an expansion of the range of structures that are explored. SHAPES found that of the five structure generators evaluated, all of them under-sampled the range of observed structural features, including complex loop topologies and functional tertiary motifs \cite{lu2025assessing}. Predicted structures can be used to augment training datasets as has been demonstrated with ESM-IF1 and Genie 2, although they provide model-derived structural information and not the corresponding biochemical activity \cite{hsu2022learning,lin2024out}. Their use therefore requires a very careful consideration of prediction uncertainty and of the sequence and structural families that are still poorly represented. Future datasets should link sequences and structures to experimental conditions and the evidence for functional annotations. Inferred annotations should be clearly distinguishable from experimental ones, and missing annotations should not be interpreted as evidence that a function is not present. Targeted data collection for under-represented families and structural arrangements would be a useful way of addressing this. An important question is whether additional data enable models to design across unfamiliar structural and functional contexts, and which types of experimental information contribute most to such ability.\\

Reliable functional design also requires a high degree of control over the physical behavior of the generated protein. A defined structure provides only part of this description when activity depends on alternative conformations and interactions with other molecules or changes in the environment. Multi-state approaches such as ProteinGenerator and DynamicMPNN already incorporate multiple structural states into protein design \cite{lisanza2025multistate,abrudan2026multi}. A further objective is to be able to design the relative stability and accessibility of the functional states, including how these properties change in response to ligand binding or other defined conditions. This will require a much closer integration of the generative models with measurements of conformational behavior and physical models of molecular interactions. Energy-based guidance and molecular simulations could be used to identify unfavorable interactions, though their usefulness will be entirely dependent on the accuracy of the energy model and the conformational space that has been sampled. Interpretability provides another route to investigating which features of a design are actually driving it. InterPLM, for instance, identified features in the protein language model representations that were associated with biological annotations and then used feature interventions to guide sequence generation \cite{simon2025interplm}. Linking those interventions to actual mutations and experimental data could help establish whether a feature is in fact providing useful control over a protein property. It is important to note that an explanation of the model behavior alone does not establish the biological mechanism responsible for the resulting activity.\\

Experimental feedback should have a bigger role in determining what models generate. The evaluation issues outlined above become particularly acute when the predicted properties are being used as the objective for optimization. A candidate can be given a better score even if its measured performance remains uncertain. ProteinGuide is one example of an approach which uses auxiliary property information in generation and then combines it with experimental data \cite{xiong2026property}. Future work in this area should examine how to select informative candidates when experimental data is sparse while taking into account both performance and uncertainty as well as diversity. It would be useful to retain both the successful and the unsuccessful designs in this process, as long as the measurements can differentiate between a design that simply fails to express and one that has lost stability or binding activity. The optimization process must also take into account the application for which the design is intended, with multiple properties potentially needing to be satisfied at the same time. In a recent study of designed binders incorporated into chimeric antigen receptors, unwanted signaling, restricted epitope accessibility, and off-target activity limited their utility \cite{chow2026sequence}. This finding necessitates introducing the relevant biological context much earlier in the process. Future workflows should make trade-offs between activity, specificity, stability, and expression, and test whether improvements persist when the design is moved beyond the assay used to guide the optimization process.\\

The practical utility of these approaches will also depend on the resources required to develop and deploy them. Faster backbone sampling, as in FrameFlow, addresses one component of such cost \cite{yim2023fast}. However, sequence design, repeated structure prediction, refinement, and experimental screening also contribute to the overall workload. Efficiency needs to be considered across the entire workflow, including the resources required to obtain experimentally useful candidates. Model compression, fewer sampling steps, and selective use of expensive calculations are directions worth exploring, with a focus on any loss of diversity or performance on difficult targets. Accessible implementations and documented training datasets would make these developments easier to reproduce and extend. Broader access should be accompanied by safeguards that are tested against the capabilities of the models being released. A study of generative protein redesign identified weaknesses in nucleic acid synthesis screening and demonstrated that updated screening methods improved detection \cite{wittmann2025strengthening}. Ongoing collaboration between model developers and experimental researchers will be crucial to evaluate these safeguards as design capabilities of the models evolve.

\section*{Competing interests}\hfill

The authors declare no conflict of interest.

\section*{Ethics declaration}\hfill

Not applicable

\section*{Acknowledgments}\hfill

This work is supported by scholarship from the Faculty of Engineering, University of Sydney to S.H.W. Figures 1 and 3 were created in BioRender. Wanasekara, S. H. (2026) https://BioRender.com/glosj6a. The authors used the OpenAI GPT model to assist with grammar and sentence structure and to improve writing during manuscript preparation. The authors reviewed and assume responsibility for the content of the article. The authors thank intellectual feedback from colleagues at the University of Sydney.


\section*{References}
\bibliography{Article}

@book{alberts2015essential,
  title={Essential cell biology},
  author={Alberts, Bruce and Bray, Dennis and Hopkin, Karen and Johnson, Alexander D and Lewis, Julian and Raff, Martin and Roberts, Keith and Walter, Peter},
  year={2015},
  publisher={Garland science}
}

@article{anfinsen1973principles,
  title={Principles that govern the folding of protein chains},
  author={Anfinsen, Christian B},
  journal={Science},
  volume={181},
  number={4096},
  pages={223--230},
  year={1973},
  publisher={American Association for the Advancement of Science}
}

@article{huang2016coming,
  title={The coming of age of de novo protein design},
  author={Huang, Po-Ssu and Boyken, Scott E and Baker, David},
  journal={Nature},
  volume={537},
  number={7620},
  pages={320--327},
  year={2016},
  publisher={Nature Publishing Group UK London}
}

@article{kuhlman2019advances,
  title={Advances in protein structure prediction and design},
  author={Kuhlman, Brian and Bradley, Philip},
  journal={Nature reviews molecular cell biology},
  volume={20},
  number={11},
  pages={681--697},
  year={2019},
  publisher={Nature Publishing Group UK London}
}

@article{bennett2026atomically,
  title={Atomically accurate de novo design of antibodies with RFdiffusion},
  author={Bennett, Nathaniel R and Watson, Joseph L and Ragotte, Robert J and Borst, Andrew J and See, D{\'e}Jena{\'e} L and Weidle, Connor and Biswas, Riti and Yu, Yutong and Shrock, Ellen L and Ault, Russell and others},
  journal={Nature},
  volume={649},
  number={8095},
  pages={183--193},
  year={2026},
  publisher={Nature Publishing Group UK London}
}

@article{berman2002protein,
  title={The protein data bank},
  author={Berman, Helen M and Battistuz, Tammy and Bhat, Talapady N and Bluhm, Wolfgang F and Bourne, Philip E and Burkhardt, Kyle and Feng, Zukang and Gilliland, Gary L and Iype, Lisa and Jain, Shri and others},
  journal={Biological Crystallography},
  volume={58},
  number={6},
  pages={899--907},
  year={2002},
  publisher={International Union of Crystallography}
}

@inproceedings{NIPS2014_f033ed80,
 author = {Goodfellow, Ian J. and Pouget-Abadie, Jean and Mirza, Mehdi and Xu, Bing and Warde-Farley, David and Ozair, Sherjil and Courville, Aaron and Bengio, Yoshua},
 booktitle = {Advances in Neural Information Processing Systems},
 editor = {Z. Ghahramani and M. Welling and C. Cortes and N. Lawrence and K. Weinberger},
 publisher = {Curran Associates, Inc.},
 title = {Generative Adversarial Nets},
 volume = {27},
 year = {2014}
}

@inproceedings{kingma2014auto,
  title={Auto-Encoding Variational Bayes},
  author={Kingma, Diederik P and Welling, Max},
  booktitle={International Conference on Learning Representations (ICLR)},
  year={2014}
}

@article{dauparas2022robust,
  title={Robust deep learning--based protein sequence design using ProteinMPNN},
  author={Dauparas, Justas and Anishchenko, Ivan and Bennett, Nathaniel and Bai, Hua and Ragotte, Robert J and Milles, Lukas F and Wicky, Basile IM and Courbet, Alexis and de Haas, Rob J and Bethel, Neville and others},
  journal={Science},
  volume={378},
  number={6615},
  pages={49--56},
  year={2022},
  publisher={American Association for the Advancement of Science}
}

@article{senior2020improved,
  title={Improved protein structure prediction using potentials from deep learning},
  author={Senior, Andrew W and Evans, Richard and Jumper, John and Kirkpatrick, James and Sifre, Laurent and Green, Tim and Qin, Chongli and {\v{Z}}{\'\i}dek, Augustin and Nelson, Alexander WR and Bridgland, Alex and others},
  journal={Nature},
  volume={577},
  number={7792},
  pages={706--710},
  year={2020},
  publisher={Nature Publishing Group UK London}
}

@article{kuhlman2003design,
  title={Design of a novel globular protein fold with atomic-level accuracy},
  author={Kuhlman, Brian and Dantas, Gautam and Ireton, Gregory C and Varani, Gabriele and Stoddard, Barry L and Baker, David},
  journal={science},
  volume={302},
  number={5649},
  pages={1364--1368},
  year={2003},
  publisher={American Association for the Advancement of Science}
}

@article{karplus2002molecular,
  title={Molecular dynamics simulations of biomolecules},
  author={Karplus, Martin and McCammon, J Andrew},
  journal={Nature structural biology},
  volume={9},
  number={9},
  pages={646--652},
  year={2002},
  publisher={Nature Publishing Group US New York}
}

@article{sugita1999replica,
  title={Replica-exchange molecular dynamics method for protein folding},
  author={Sugita, Yuji and Okamoto, Yuko},
  journal={Chemical physics letters},
  volume={314},
  number={1-2},
  pages={141--151},
  year={1999},
  publisher={Elsevier}
}

@article{hollingsworth2018molecular,
  title={Molecular dynamics simulation for all},
  author={Hollingsworth, Scott A and Dror, Ron O},
  journal={Neuron},
  volume={99},
  number={6},
  pages={1129--1143},
  year={2018},
  publisher={Elsevier}
}

@incollection{leaver2011rosetta3,
  title={ROSETTA3: an object-oriented software suite for the simulation and design of macromolecules},
  author={Leaver-Fay, Andrew and Tyka, Michael and Lewis, Steven M and Lange, Oliver F and Thompson, James and Jacak, Ron and Kaufman, Kristian W and Renfrew, P Douglas and Smith, Colin A and Sheffler, Will and others},
  booktitle={Methods in enzymology},
  volume={487},
  pages={545--574},
  year={2011},
  publisher={Elsevier}
}

@article{ching2018opportunities,
  title={Opportunities and obstacles for deep learning in biology and medicine},
  author={Ching, Travers and Himmelstein, Daniel S and Beaulieu-Jones, Brett K and Kalinin, Alexandr A and Do, Brian T and Way, Gregory P and Ferrero, Enrico and Agapow, Paul-Michael and Zietz, Michael and Hoffman, Michael M and others},
  journal={Journal of the royal society interface},
  volume={15},
  number={141},
  pages={20170387},
  year={2018},
  publisher={The Royal Society}
}

@article{jumper2021highly,
  title={Highly accurate protein structure prediction with AlphaFold},
  author={Jumper, John and Evans, Richard and Pritzel, Alexander and Green, Tim and Figurnov, Michael and Ronneberger, Olaf and Tunyasuvunakool, Kathryn and Bates, Russ and {\v{Z}}{\'\i}dek, Augustin and Potapenko, Anna and others},
  journal={nature},
  volume={596},
  number={7873},
  pages={583--589},
  year={2021},
  publisher={Nature Publishing Group UK London}
}

@article{zhou2015predicting,
  title={Predicting effects of noncoding variants with deep learning--based sequence model},
  author={Zhou, Jian and Troyanskaya, Olga G},
  journal={Nature methods},
  volume={12},
  number={10},
  pages={931--934},
  year={2015},
  publisher={Nature Publishing Group US New York}
}

@article{lin2023evolutionary,
  title={Evolutionary-scale prediction of atomic-level protein structure with a language model},
  author={Lin, Zeming and Akin, Halil and Rao, Roshan and Hie, Brian and Zhu, Zhongkai and Lu, Wenting and Smetanin, Nikita and Verkuil, Robert and Kabeli, Ori and Shmueli, Yaniv and others},
  journal={Science},
  volume={379},
  number={6637},
  pages={1123--1130},
  year={2023},
  publisher={American Association for the Advancement of Science}
}

@article{wagle2026interpretable,
  title={Interpretable deep generative ensemble learning for single-cell omics with Hydra},
  author={Wagle, Manoj M and Liu, Chunlei and Liu, Zunpeng and Wang, Yongheng and Kellis, Manolis and Patrick, Ellis and Yang, Pengyi},
  journal={Molecular Systems Biology},
  pages={1--19},
  year={2026},
  publisher={Springer}
}

@article{litjens2017survey,
  title={A survey on deep learning in medical image analysis},
  author={Litjens, Geert and Kooi, Thijs and Bejnordi, Babak Ehteshami and Setio, Arnaud Arindra Adiyoso and Ciompi, Francesco and Ghafoorian, Mohsen and Van Der Laak, Jeroen Awm and Van Ginneken, Bram and S{\'a}nchez, Clara I},
  journal={Medical image analysis},
  volume={42},
  pages={60--88},
  year={2017},
  publisher={Elsevier}
}

@article{baek2021accurate,
  title={Accurate prediction of protein structures and interactions using a three-track neural network},
  author={Baek, Minkyung and DiMaio, Frank and Anishchenko, Ivan and Dauparas, Justas and Ovchinnikov, Sergey and Lee, Gyu Rie and Wang, Jue and Cong, Qian and Kinch, Lisa N and Schaeffer, R Dustin and others},
  journal={Science},
  volume={373},
  number={6557},
  pages={871--876},
  year={2021},
  publisher={American Association for the Advancement of Science}
}

@article{rives2021biological,
  title={Biological structure and function emerge from scaling unsupervised learning to 250 million protein sequences},
  author={Rives, Alexander and Meier, Joshua and Sercu, Tom and Goyal, Siddharth and Lin, Zeming and Liu, Jason and Guo, Demi and Ott, Myle and Zitnick, C Lawrence and Ma, Jerry and others},
  journal={Proceedings of the national academy of sciences},
  volume={118},
  number={15},
  pages={e2016239118},
  year={2021},
  publisher={National Academy of Sciences}
}

@article{meier2021language,
  title={Language models enable zero-shot prediction of the effects of mutations on protein function},
  author={Meier, Joshua and Rao, Roshan and Verkuil, Robert and Liu, Jason and Sercu, Tom and Rives, Alex},
  journal={Advances in neural information processing systems},
  volume={34},
  pages={29287--29303},
  year={2021}
}

@article{nijkamp2023progen2,
  title={Progen2: exploring the boundaries of protein language models},
  author={Nijkamp, Erik and Ruffolo, Jeffrey A and Weinstein, Eli N and Naik, Nikhil and Madani, Ali},
  journal={Cell systems},
  volume={14},
  number={11},
  pages={968--978},
  year={2023},
  publisher={Elsevier}
}

@article{watson2023novo,
  title={De novo design of protein structure and function with RFdiffusion},
  author={Watson, Joseph L and Juergens, David and Bennett, Nathaniel R and Trippe, Brian L and Yim, Jason and Eisenach, Helen E and Ahern, Woody and Borst, Andrew J and Ragotte, Robert J and Milles, Lukas F and others},
  journal={Nature},
  volume={620},
  number={7976},
  pages={1089--1100},
  year={2023},
  publisher={Nature Publishing Group UK London}
}

@article{yim2023fast,
  title={Fast protein backbone generation with se (3) flow matching},
  author={Yim, Jason and Campbell, Andrew and Foong, Andrew YK and Gastegger, Michael and Jim{\'e}nez-Luna, Jos{\'e} and Lewis, Sarah and Satorras, Victor Garcia and Veeling, Bastiaan S and Barzilay, Regina and Jaakkola, Tommi and others},
  journal={arXiv preprint arXiv:2310.05297},
  year={2023}
}

@article{abramson2024accurate,
  title={Accurate structure prediction of biomolecular interactions with AlphaFold 3},
  author={Abramson, Josh and Adler, Jonas and Dunger, Jack and Evans, Richard and Green, Tim and Pritzel, Alexander and Ronneberger, Olaf and Willmore, Lindsay and Ballard, Andrew J and Bambrick, Joshua and others},
  journal={Nature},
  volume={630},
  number={8016},
  pages={493--500},
  year={2024},
  publisher={Nature Publishing Group UK London}
}

@article{hayes2025simulating,
  title={Simulating 500 million years of evolution with a language model},
  author={Hayes, Thomas and Rao, Roshan and Akin, Halil and Sofroniew, Nicholas J and Oktay, Deniz and Lin, Zeming and Verkuil, Robert and Tran, Vincent Q and Deaton, Jonathan and Wiggert, Marius and others},
  journal={Science},
  volume={387},
  number={6736},
  pages={850--858},
  year={2025},
  publisher={American Association for the Advancement of Science}
}

@article{gulrajani2017improved,
  title={Improved training of wasserstein gans},
  author={Gulrajani, Ishaan and Ahmed, Faruk and Arjovsky, Martin and Dumoulin, Vincent and Courville, Aaron},
  journal={Advances in neural information processing systems},
  volume={30},
  year={2017}
}

@article{ho2020denoising,
  title={Denoising diffusion probabilistic models},
  author={Ho, Jonathan and Jain, Ajay and Abbeel, Pieter},
  journal={Advances in neural information processing systems},
  volume={33},
  pages={6840--6851},
  year={2020}
}

@inproceedings{lipman2023flow,
  title = {Flow Matching for Generative Modeling},
  author={Lipman, Yaron and Chen, Ricky TQ and Ben-Hamu, Heli and Nickel, Maximilian and Le, Matt},
  booktitle = {International Conference on Learning Representations (ICLR)},
  year      = {2023}
}

@article{vaswani2017attention,
  title={Attention is all you need},
  author={Vaswani, Ashish and Shazeer, Noam and Parmar, Niki and Uszkoreit, Jakob and Jones, Llion and Gomez, Aidan N and Kaiser, {\L}ukasz and Polosukhin, Illia},
  journal={Advances in neural information processing systems},
  volume={30},
  year={2017}
}

@article{du2021trrosetta,
  title={The trRosetta server for fast and accurate protein structure prediction},
  author={Du, Zongyang and Su, Hong and Wang, Wenkai and Ye, Lisha and Wei, Hong and Peng, Zhenling and Anishchenko, Ivan and Baker, David and Yang, Jianyi},
  journal={Nature protocols},
  volume={16},
  number={12},
  pages={5634--5651},
  year={2021},
  publisher={Nature Publishing Group UK London}
}

@article{wu2024protein,
  title={Protein structure generation via folding diffusion},
  author={Wu, Kevin E and Yang, Kevin K and Van den Berg, Rianne and Alamdari, Sarah and Zou, James Y and Lu, Alex X and Amini, Ava P},
  journal={Nature communications},
  volume={15},
  number={1},
  pages={1059},
  year={2024},
  publisher={Nature Publishing Group UK London}
}

@inproceedings{
    jing2021learning,
    title={Learning from Protein Structure with Geometric Vector Perceptrons},
    author={Bowen Jing and Stephan Eismann and Patricia Suriana and Raphael John Lamarre Townshend and Ron Dror},
    booktitle={International Conference on Learning Representations},
    year={2021}
}

@inproceedings{rao2021msa,
  title={MSA transformer},
  author={Rao, Roshan M and Liu, Jason and Verkuil, Robert and Meier, Joshua and Canny, John and Abbeel, Pieter and Sercu, Tom and Rives, Alexander},
  booktitle={International conference on machine learning},
  pages={8844--8856},
  year={2021},
  organization={PMLR}
}

@article{radford2018improving,
  title={Improving language understanding by generative pre-training},
  author={Radford, Alec and Narasimhan, Karthik and Salimans, Tim and Sutskever, Ilya and others},
  year={2018},
  publisher={San Francisco, CA, USA}
}

@article{su2025trimodal,
  title={A trimodal protein language model enables advanced protein searches},
  author={Su, Jin and He, Yan and You, Shiyang and Jiang, Shiyu and Zhou, Xibin and Zhang, Xuting and Wang, Yuxuan and Su, Xining and Tolstoy, Igor and Chang, Xing and others},
  journal={Nature Biotechnology},
  pages={1--7},
  year={2025},
  publisher={Nature Publishing Group US, New York}
}

@article{liu2025text,
  title={A text-guided protein design framework},
  author={Liu, Shengchao and Li, Yanjing and Li, Zhuoxinran and Gitter, Anthony and Zhu, Yutao and Lu, Jiarui and Xu, Zhao and Nie, Weili and Ramanathan, Arvind and Xiao, Chaowei and others},
  journal={Nature Machine Intelligence},
  volume={7},
  number={4},
  pages={580--591},
  year={2025},
  publisher={Nature Publishing Group UK, London}
}

@article{wagle2026deep,
  title={Deep interpretable learning of sample representations for characterizing disease states in single-cell transcriptomics},
  author={Wagle, Manoj M and Wang, Yongheng and Samanta, Soham and Liu, Zunpeng and Patrick, Ellis and Yang, Pengyi and Kellis, Manolis},
  journal={bioRxiv},
  pages={2026--07},
  year={2026},
  publisher={Cold Spring Harbor Laboratory}
}

@article{hawkins2021generating,
  title={Generating functional protein variants with variational autoencoders},
  author={Hawkins-Hooker, Alex and Depardieu, Florence and Baur, Sebastien and Couairon, Guillaume and Chen, Arthur and Bikard, David},
  journal={PLoS computational biology},
  volume={17},
  number={2},
  pages={e1008736},
  year={2021},
  publisher={Public Library of Science, San Francisco, CA USA}
}

@article{repecka2021expanding,
  title={Expanding functional protein sequence spaces using generative adversarial networks},
  author={Repecka, Donatas and Jauniskis, Vykintas and Karpus, Laurynas and Rembeza, Elzbieta and Rokaitis, Irmantas and Zrimec, Jan and Poviloniene, Simona and Laurynenas, Audrius and Viknander, Sandra and Abuajwa, Wissam and others},
  journal={Nature Machine Intelligence},
  volume={3},
  number={4},
  pages={324--333},
  year={2021},
  publisher={Nature Publishing Group UK, London}
}

@article{truong2023poet,
  title={Poet: A generative model of protein families as sequences-of-sequences},
  author={Truong Jr, Timothy and Bepler, Tristan},
  journal={Advances in Neural Information Processing Systems},
  volume={36},
  pages={77379--77415},
  year={2023}
}

@misc{ferruzprotgpt2,
  title={ProtGPT2 is a deep unsupervised language model for protein design. Nat Commun. 2022; 13 (1): 4348},
  author={Ferruz, N and Schmidt, S and H{\"o}cker, B},
  publisher={Epub 2022/07/28. doi: 10.1038/s41467-022-32007-7. PubMed PMID: 35896542}
}

@article{bhatnagar2026scaling,
  title={Scaling unlocks broader generation and deeper functional understanding of proteins},
  author={Bhatnagar, Aadyot and Jain, Sarthak and Beazer, Joel and Curran, Samuel and Hoffnagle, Alexander and Ching, Kyle and Martyn, Michael and Nayfach, Stephen and Ruffolo, Jeffrey and Madani, Ali},
  journal={Advances in Neural Information Processing Systems},
  volume={38},
  pages={46109--46145},
  year={2026}
}

@article{alamdari2023protein,
  title={Protein generation with evolutionary diffusion: sequence is all you need},
  author={Alamdari, Sarah and Thakkar, Nitya and Van Den Berg, Rianne and Tenenholtz, Neil and Strome, Robert and Moses, Alan M and Lu, Alex X and Fusi, Nicol{\`o} and Amini, Ava P and Yang, Kevin K},
  journal={BioRxiv},
  pages={2023--09},
  year={2023},
  publisher={Cold Spring Harbor Laboratory}
}

@article{wang2024diffusion,
  title={Diffusion language models are versatile protein learners},
  author={Wang, Xinyou and Zheng, Zaixiang and Ye, Fei and Xue, Dongyu and Huang, Shujian and Gu, Quanquan},
  journal={arXiv preprint arXiv:2402.18567},
  year={2024}
}

@article{meshchaninov2024diffusion,
  title={Diffusion on language model encodings for protein sequence generation},
  author={Meshchaninov, Viacheslav and Strashnov, Pavel and Shevtsov, Andrey and Nikolaev, Fedor and Ivanisenko, Nikita and Kardymon, Olga and Vetrov, Dmitry},
  journal={arXiv preprint arXiv:2403.03726},
  year={2024}
}

@article{atkinson2025protein,
  title={Protein sequence modelling with Bayesian flow networks},
  author={Atkinson, Timothy and Barrett, Thomas D and Cameron, Scott and Guloglu, Bora and Greenig, Matthew and Tan, Charlie B and Robinson, Louis and Graves, Alex and Copoiu, Liviu and Laterre, Alexandre},
  journal={Nature Communications},
  volume={16},
  number={1},
  pages={3197},
  year={2025},
  publisher={Nature Publishing Group UK, London}
}

@article{madani2023large,
  title={Large language models generate functional protein sequences across diverse families},
  author={Madani, Ali and Krause, Ben and Greene, Eric R and Subramanian, Subu and Mohr, Benjamin P and Holton, James M and Olmos Jr, Jose Luis and Xiong, Caiming and Sun, Zachary Z and Socher, Richard and others},
  journal={Nature biotechnology},
  volume={41},
  number={8},
  pages={1099--1106},
  year={2023},
  publisher={Nature Publishing Group US, New York}
}

@inproceedings{munsamy2022zymctrl,
  title={ZymCTRL: a conditional language model for the controllable generation of artificial enzymes},
  author={Munsamy, Geraldene and Lindner, Sebastian and Lorenz, Philipp and Ferruz, Noelia},
  booktitle={NeurIPS machine learning in structural biology workshop},
  year={2022},
  organization={NeurIPS}
}

@article{kucera2022conditional,
  title={Conditional generative modeling for de novo protein design with hierarchical functions},
  author={Kucera, Tim and Togninalli, Matteo and Meng-Papaxanthos, Laetitia},
  journal={Bioinformatics},
  volume={38},
  number={13},
  pages={3454--3461},
  year={2022},
  publisher={Oxford University Press}
}

@article{lee2023score,
  title={Score-based generative modeling for de novo protein design},
  author={Lee, Jin Sub and Kim, Jisun and Kim, Philip M},
  journal={Nature Computational Science},
  volume={3},
  number={5},
  pages={382--392},
  year={2023},
  publisher={Nature Publishing Group US, New York}
}

@article{trippe2022diffusion,
  title={Diffusion probabilistic modeling of protein backbones in 3d for the motif-scaffolding problem},
  author={Trippe, Brian L and Yim, Jason and Tischer, Doug and Baker, David and Broderick, Tamara and Barzilay, Regina and Jaakkola, Tommi},
  journal={arXiv preprint arXiv:2206.04119},
  year={2022}
}

@article{lin2023generating,
  title={Generating novel, designable, and diverse protein structures by equivariantly diffusing oriented residue clouds},
  author={Lin, Yeqing and AlQuraishi, Mohammed},
  journal={arXiv preprint arXiv:2301.12485},
  year={2023}
}

@article{yim2023se,
  title={SE (3) diffusion model with application to protein backbone generation},
  author={Yim, Jason and Trippe, Brian L and De Bortoli, Valentin and Mathieu, Emile and Doucet, Arnaud and Barzilay, Regina and Jaakkola, Tommi},
  journal={arXiv preprint arXiv:2302.02277},
  year={2023}
}

@article{ingraham2023illuminating,
  title={Illuminating protein space with a programmable generative model},
  author={Ingraham, John B and Baranov, Max and Costello, Zak and Barber, Karl W and Wang, Wujie and Ismail, Ahmed and Frappier, Vincent and Lord, Dana M and Ng-Thow-Hing, Christopher and Van Vlack, Erik R and others},
  journal={Nature},
  volume={623},
  number={7989},
  pages={1070--1078},
  year={2023},
  publisher={Nature Publishing Group UK, London}
}

@article{butcher2026atomic,
  title={Atomic functional site scaffolding in biopolymers with RFdiffusion3},
  author={Butcher, Jasper and Krishna, Rohith and Mitra, Raktim and Woodbury, Seth M and Kim, Donghyo and Brent, Rafael I and Li, Yanjing and Corley, Nathaniel and Favor, Andrew and Chen, Anqi and others},
  journal={Science},
  pages={eaeh2982},
  year={2026},
  publisher={American Association for the Advancement of Science}
}

@inproceedings{bose2024se,
  title={Se (3)-stochastic flow matching for protein backbone generation},
  author={Bose, Joey and Akhound-Sadegh, Tara and Huguet, Guillaume and Fatras, Kilian and Rector-Brooks, Jarrid and Liu, Chenghao and Nica, Andrei and Korablyov, Maksym and Bronstein, Michael and Tong, Alexander},
  booktitle={International Conference on Learning Representations},
  volume={2024},
  pages={22590--22621},
  year={2024}
}

@article{huguet2024sequence,
  title={Sequence-augmented SE (3)-flow matching for conditional protein generation},
  author={Huguet, Guillaume and Vuckovic, James and Fatras, Kilian and Thibodeau-Laufer, Eric and Lemos, Pablo and Islam, Riashat and Liu, Cheng-Hao and Rector-Brooks, Jarrid and Akhound-Sadegh, Tara and Bronstein, Michael and others},
  journal={Advances in neural information processing systems},
  volume={37},
  pages={33007--33036},
  year={2024}
}

@article{lin2024out,
  title={Out of many, one: Designing and scaffolding proteins at the scale of the structural universe with genie 2},
  author={Lin, Yeqing and Lee, Minji and Zhang, Zhao and AlQuraishi, Mohammed},
  journal={arXiv preprint arXiv:2405.15489},
  year={2024}
}

@inproceedings{geffner2025proteina,
  title={Proteina: Scaling flow-based protein structure generative models},
  author={Geffner, Tomas and Didi, Kieran and Zhang, Zuobai and Reidenbach, Danny and Cao, Zhonglin and Yim, Jason and Geiger, Mario and Dallago, Christian and Kucukbenli, Emine and Vahdat, Arash and others},
  booktitle={International Conference on Learning Representations},
  volume={2025},
  pages={98803--98851},
  year={2025}
}

@inproceedings{hsu2022learning,
  title={Learning inverse folding from millions of predicted structures},
  author={Hsu, Chloe and Verkuil, Robert and Liu, Jason and Lin, Zeming and Hie, Brian and Sercu, Tom and Lerer, Adam and Rives, Alexander},
  booktitle={International conference on machine learning},
  pages={8946--8970},
  year={2022},
  organization={PMLR}
}

@article{ren2024accurate,
  title={Accurate and robust protein sequence design with CarbonDesign},
  author={Ren, Milong and Yu, Chungong and Bu, Dongbo and Zhang, Haicang},
  journal={Nature Machine Intelligence},
  volume={6},
  number={5},
  pages={536--547},
  year={2024},
  publisher={Nature Publishing Group UK, London}
}

@inproceedings{zheng2023structure,
  title={Structure-informed language models are protein designers},
  author={Zheng, Zaixiang and Deng, Yifan and Xue, Dongyu and Zhou, Yi and Ye, Fei and Gu, Quanquan},
  booktitle={International conference on machine learning},
  pages={42317--42338},
  year={2023},
  organization={PMLR}
}

@article{karimi2020novo,
  title={De novo protein design for novel folds using guided conditional wasserstein generative adversarial networks},
  author={Karimi, Mostafa and Zhu, Shaowen and Cao, Yue and Shen, Yang},
  journal={Journal of chemical information and modeling},
  volume={60},
  number={12},
  pages={5667--5681},
  year={2020},
  publisher={ACS Publications}
}

@article{goverde2023novo,
  title={De novo protein design by inversion of the AlphaFold structure prediction network},
  author={Goverde, Casper A and Wolf, Benedict and Khakzad, Hamed and Rosset, St{\'e}phane and Correia, Bruno E},
  journal={Protein Science},
  volume={32},
  number={6},
  pages={e4653},
  year={2023},
  publisher={Wiley Online Library}
}

@article{dauparas2025atomic,
  title={Atomic context-conditioned protein sequence design using LigandMPNN},
  author={Dauparas, Justas and Lee, Gyu Rie and Pecoraro, Robert and An, Linna and Anishchenko, Ivan and Glasscock, Cameron and Baker, David},
  journal={Nature Methods},
  volume={22},
  number={4},
  pages={717--723},
  year={2025},
  publisher={Nature Publishing Group US, New York}
}

@article{yi2025all,
  title={All-atom inverse protein folding through discrete flow matching},
  author={Yi, Kai and Jamali, Kiarash and Scheres, Sjors HW},
  journal={arXiv preprint arXiv:2507.14156},
  year={2025}
}

@inproceedings{abrudan2026multi,
  title={Multi-state Protein Sequence Design with DynamicMPNN},
  author={Abrudan, Alex and Pujalte Ojeda, Sebastian and Joshi, Chaitanya and Greenig, Matthew and Engelberger, Felipe and Khmelinskaia, Alena and Meiler, Jens and Vendruscolo, Michele and Knowles, Tuomas},
  booktitle={International Conference on Learning Representations},
  volume={2026},
  pages={153620--153641},
  year={2026}
}

@article{lisanza2025multistate,
  title={Multistate and functional protein design using RoseTTAFold sequence space diffusion},
  author={Lisanza, Sidney Lyayuga and Gershon, Jacob Merle and Tipps, Samuel WK and Sims, Jeremiah Nelson and Arnoldt, Lucas and Hendel, Samuel J and Simma, Miriam K and Liu, Ge and Yase, Muna and Wu, Hongwei and others},
  journal={Nature biotechnology},
  volume={43},
  number={8},
  pages={1288--1298},
  year={2025},
  publisher={Nature Publishing Group US, New York}
}

@inproceedings{ren2024carbonnovo,
  title={Carbonnovo: Joint design of protein structure and sequence using a unified energy-based model},
  author={Ren, Milong and Zhu, Tian and Zhang, Haicang},
  booktitle={Forty-first international conference on machine learning},
  year={2024}
}

@article{campbell2024generative,
  title={Generative flows on discrete state-spaces: Enabling multimodal flows with applications to protein co-design},
  author={Campbell, Andrew and Yim, Jason and Barzilay, Regina and Rainforth, Tom and Jaakkola, Tommi},
  journal={arXiv preprint arXiv:2402.04997},
  year={2024}
}

@article{chu2024all,
  title={An all-atom protein generative model},
  author={Chu, Alexander E and Kim, Jinho and Cheng, Lucy and El Nesr, Gina and Xu, Minkai and Shuai, Richard W and Huang, Po-Ssu},
  journal={Proceedings of the National Academy of Sciences},
  volume={121},
  number={27},
  pages={e2311500121},
  year={2024},
  publisher={National Academy of Sciences}
}

@inproceedings{geffner2026proteina,
  title={La-proteina: Atomistic protein generation via partially latent flow matching},
  author={Geffner, Tomas and Didi, Kieran and Cao, Zhonglin and Reidenbach, Danny and Zhang, Zuobai and Dallago, Christian and Kucukbenli, Emine and Kreis, Karsten and Vahdat, Arash},
  booktitle={International Conference on Learning Representations},
  volume={2026},
  pages={67350--67395},
  year={2026}
}

@article{reidenbach2025consistent,
  title={Consistent synthetic sequences unlock structural diversity in fully atomistic de novo protein design},
  author={Reidenbach, Danny and Cao, Zhonglin and Zhang, Zuobai and Didi, Kieran and Geffner, Tomas and Zhou, Guoqing and Tang, Jian and Dallago, Christian and Vahdat, Arash and Kucukbenli, Emine and others},
  journal={arXiv preprint arXiv:2512.01976},
  year={2025}
}

@inproceedings{wang2025dplm,
  title={Dplm-2: A multimodal diffusion protein language model},
  author={Wang, Xinyou and Zheng, Zaixiang and Ye, Fei and Xue, Dongyu and Huang, Shujian and Gu, Quanquan},
  booktitle={International Conference on Learning Representations},
  volume={2025},
  pages={35463--35490},
  year={2025}
}

@article{yang2025co,
  title={Co-design protein sequence and structure in discrete space via generative flow},
  author={Yang, Sen and Ju, Lingli and Cheng, Peng and Zhou, JiangLin and Cai, Yamin and Feng, Dawei},
  journal={Bioinformatics},
  volume={41},
  number={5},
  pages={btaf248},
  year={2025},
  publisher={Oxford University Press}
}

@article{chen2026co,
  title={Co-Generative De Novo Functional Protein Design},
  author={Chen, Xinrui and Luo, Yizhen and Fan, Siqi and Nie, Zaiqing},
  journal={arXiv preprint arXiv:2605.00948},
  year={2026}
}

@article{yim2024improved,
  title={Improved motif-scaffolding with SE (3) flow matching},
  author={Yim, Jason and Campbell, Andrew and Mathieu, Emile and Foong, Andrew YK and Gastegger, Michael and Jim{\'e}nez-Luna, Jos{\'e} and Lewis, Sarah and Satorras, Victor Garcia and Veeling, Bastiaan S and No{\'e}, Frank and others},
  journal={ArXiv},
  pages={arXiv--2401},
  year={2024}
}

@article{qu2026protein,
  title={Protein Autoregressive Modeling via Multiscale Structure Generation},
  author={Qu, Yanru and Hsieh, Cheng-Yen and Zheng, Zaixiang and Liu, Ge and Gu, Quanquan},
  journal={arXiv preprint arXiv:2602.04883},
  year={2026}
}

@article{braun2026computational,
  title={Computational enzyme design by catalytic motif scaffolding},
  author={Braun, Markus and Tripp, Adrian and Chakatok, Morakot and Kaltenbrunner, Sigrid and Fischer, Celina and Stoll, David and Bijelic, Aleksandar and Elaily, Wael and Totaro, Massimo G and Moser, Melanie and others},
  journal={Nature},
  volume={649},
  number={8095},
  pages={237--245},
  year={2026},
  publisher={Nature Publishing Group UK, London}
}

@article{ahern2026atom,
  title={Atom-level enzyme active site scaffolding using RFdiffusion2},
  author={Ahern, Woody and Yim, Jason and Tischer, Doug and Salike, Saman and Woodbury, Seth M and Kim, Donghyo and Kalvet, Indrek and Kipnis, Yakov and Coventry, Brian and Altae-Tran, Han Raut and others},
  journal={Nature Methods},
  volume={23},
  number={1},
  pages={96--105},
  year={2026},
  publisher={Nature Publishing Group US, New York}
}

@article{qian2026complex,
  title={Complex enzyme active site scaffolding by iteratively detuned catalytic guidance},
  author={Qian, Tingli and Wang, Weijin and Gao, Shilong and Muraishi, Aiko and Zhao, Beibei and An, Davi Nakajima and Xue, Fanglei and Kazman, Max and Mendoza, Johnny and Nguyen, Hannah and others},
  journal={bioRxiv},
  pages={2026--10},
  year={2026},
  publisher={Cold Spring Harbor Laboratory}
}

@article{pacesa2025one,
  title={One-shot design of functional protein binders with BindCraft},
  author={Pacesa, Martin and Nickel, Lennart and Schellhaas, Christian and Schmidt, Joseph and Pyatova, Ekaterina and Kissling, Lucas and Barendse, Patrick and Choudhury, Jagrity and Kapoor, Srajan and Alcaraz-Serna, Ana and others},
  journal={Nature},
  volume={646},
  number={8084},
  pages={483--492},
  year={2025},
  publisher={Nature Publishing Group UK, London}
}

@inproceedings{didi2026scaling,
  title={Scaling atomistic protein binder design with generative pretraining and test-time compute},
  author={Didi, Kieran and Zhang, Zuobai and Zhou, Guoqing and Reidenbach, Danny and Cao, Zhonglin and Cha, Sooyoung and Geffner, Tomas and Dallago, Christian and Tang, Jian and Bronstein, Michael and others},
  booktitle={International Conference on Learning Representations},
  volume={2026},
  pages={61056--61109},
  year={2026}
}

@article{didi2026latent,
  title={Latent generative search unlocks de novo design of untapped biomolecular interactions at scale.},
  author={Didi, Kieran and Reidenbach, Danny and Penner, Matthew and Ravichandran, Supriya and Case, Marshall and Nichols, Mike and Swanson, Erik and Reis, Alex and Prescott, Maggie and Qian, Yue and others},
  journal={Biorxiv: the Preprint Server for Biology},
  year={2026}
}

@article{jin2021iterative,
  title={Iterative refinement graph neural network for antibody sequence-structure co-design},
  author={Jin, Wengong and Wohlwend, Jeremy and Barzilay, Regina and Jaakkola, Tommi},
  journal={arXiv preprint arXiv:2110.04624},
  year={2021}
}

@article{luo2022antigen,
  title={Antigen-specific antibody design and optimization with diffusion-based generative models for protein structures},
  author={Luo, Shitong and Su, Yufeng and Peng, Xingang and Wang, Sheng and Peng, Jian and Ma, Jianzhu},
  journal={Advances in Neural Information Processing Systems},
  volume={35},
  pages={9754--9767},
  year={2022}
}

@article{mille2026efficient,
  title={Efficient generation of epitope-targeted antibodies with Germinal},
  author={Mille-Fragoso, Luis S and Driscoll, Claudia L and Wang, John N and Dai, Haoyu and Widatalla, Talal and Zhang, Jim L and Zhang, Xiaowei and Rao, Bing and Feng, Liang and Hie, Brian L and others},
  journal={Nature biotechnology},
  pages={1--10},
  year={2026},
  publisher={Nature Publishing Group}
}

@article{stark2025boltzgen,
  title={Boltzgen: Toward universal binder design},
  author={Stark, Hannes and Faltings, Felix and Choi, MinGyu and Xie, Yuxin and Hur, Eunsu and O’Donnell, Timothy and Bushuiev, Anton and U{\c{c}}ar, Talip and Passaro, Saro and Mao, Weian and others},
  journal={BioRxiv},
  year={2025}
}

@article{lee2026designing,
  title={Designing Protein Structures from Natural Language},
  author={Lee, Joseph and Li, Yanjing and Qiang, Bo and Lau, Gregory and Ettinger, Allyson and Smith, Noah A and Donovan-Maiye, Rory and Baker, David and Hajishirzi, Hanna and Koh, Pang Wei},
  journal={arXiv preprint arXiv:2610.13917},
  year={2026}
}

@article{zheng2025motifbench,
  title={MotifBench: A standardized protein design benchmark for motif-scaffolding problems},
  author={Zheng, Zhuoqi and Zhang, Bo and Didi, Kieran and Yang, Kevin K and Yim, Jason and Watson, Joseph L and Chen, Hai-Feng and Trippe, Brian L},
  journal={arXiv preprint arXiv:2502.12479},
  year={2025}
}

@InProceedings{pmlr-v306-kuang26d,
  title = 	 {{PDFB}ench: A Benchmark for De Novo Protein Design from Function},
  author =       {Kuang, Jiahao and Liu, Nuowei and Sun, Changzhi and Wang, Jie and Ji, Tao and Wu, Yuanbin},
  booktitle = 	 {Proceedings of the 43rd International Conference on Machine Learning},
  pages = 	 {61349--61371},
  year = 	 {2026},
  editor = 	 {Zhang, Tong and Dudik, Miroslav and Jaggi, Martin and Agarwal, Alekh and Li, Sharon and Schuurmans, Dale and Zhu, Jerry and Berkenkamp, Felix and Dong, Hanze and Bietti, Alberto},
  volume = 	 {306},
  series = 	 {Proceedings of Machine Learning Research},
  month = 	 {06--11 Jul},
  publisher =    {PMLR},
}

@article{garcia2026evaluating,
  title={Evaluating zero-shot prediction of monomeric protein design success by AlphaFold, ESMFold, and ProteinMPNN},
  author={Garcia, Mario and Dixit, Sugyan M and Rocklin, Gabriel J},
  journal={Protein Science},
  volume={35},
  number={2},
  pages={e70453},
  year={2026},
  publisher={Wiley Online Library}
}

@article{chen2010molprobity,
  title={MolProbity: all-atom structure validation for macromolecular crystallography},
  author={Chen, Vincent B and Arendall, W Bryan and Headd, Jeffrey J and Keedy, Daniel A and Immormino, Robert M and Kapral, Gary J and Murray, Laura W and Richardson, Jane S and Richardson, David C},
  journal={Biological crystallography},
  volume={66},
  number={1},
  pages={12--21},
  year={2010},
  publisher={International Union of Crystallography}
}

@article{lu2025assessing,
  title={Assessing generative model coverage of protein structures with SHAPES},
  author={Lu, Tianyu and Liu, Melissa and Chen, Yilin and Kim, Jinho and Huang, Po-Ssu},
  journal={Cell Systems},
  volume={16},
  number={8},
  year={2025},
  publisher={Elsevier}
}

@article{simon2025interplm,
  title={InterPLM: discovering interpretable features in protein language models via sparse autoencoders},
  author={Simon, Elana and Zou, James},
  journal={Nature methods},
  volume={22},
  number={10},
  pages={2107--2117},
  year={2025},
  publisher={Nature Publishing Group US, New York}
}

@article{xiong2026property,
  title={Property guidance for protein sequence generative models with ProteinGuide},
  author={Xiong, Junhao and Gaur, Ishan and Lukarska, Maria and Nisonoff, Hunter and Oltrogge, Luke M and Savage, David F and Listgarten, Jennifer},
  journal={Nature Biotechnology},
  pages={1--13},
  year={2026},
  publisher={Nature Publishing Group US, New York}
}

@article{chow2026sequence,
  title={Sequence and structural determinants of efficacious de novo chimaeric antigen receptors},
  author={Chow, Arthur and Chu, Hoyin and Li, Ruofan and Nalbant, Benan N and Dozic, Abdul Vehab and Kida, Laura C and Tang, Zeyu and Palmeri, Joseph R and Lareau, Caleb A},
  journal={Nature biomedical engineering},
  pages={1--16},
  year={2026},
  publisher={Nature Publishing Group UK, London}
}

@article{wittmann2025strengthening,
  title={Strengthening nucleic acid biosecurity screening against generative protein design tools},
  author={Wittmann, Bruce J and Alexanian, Tessa and Bartling, Craig and Beal, Jacob and Clore, Adam and Diggans, James and Flyangolts, Kevin and Gemler, Bryan T and Mitchell, Tom and Murphy, Steven T and others},
  journal={Science},
  volume={390},
  number={6768},
  pages={82--87},
  year={2025},
  publisher={American Association for the Advancement of Science}
}

@article{kendrew1958three,
  title={A three-dimensional model of the myoglobin molecule obtained by x-ray analysis},
  author={Kendrew, John C and Bodo, G and Dintzis, Howard M and Parrish, RG and Wyckoff, Harold and Phillips, David C},
  journal={Nature},
  volume={181},
  number={4610},
  pages={662--666},
  year={1958},
  publisher={Nature Publishing Group UK, London}
}

@article{williamson1985solution,
  title={Solution conformation of proteinase inhibitor IIA from bull seminal plasma by 1H nuclear magnetic resonance and distance geometry},
  author={Williamson, Michael P and Havel, Timothy F and W{\"u}thrich, Kurt},
  journal={Journal of molecular biology},
  volume={182},
  number={2},
  pages={295--315},
  year={1985},
  publisher={Elsevier}
}

@article{bai2013ribosome,
  title={Ribosome structures to near-atomic resolution from thirty thousand cryo-EM particles},
  author={Bai, Xiao-chen and Fernandez, Israel S and McMullan, Greg and Scheres, Sjors HW},
  journal={elife},
  volume={2},
  pages={e00461},
  year={2013},
  publisher={eLife Sciences Publications, Ltd}
}

@article{anfinsen1961kinetics,
  title={The kinetics of formation of native ribonuclease during oxidation of the reduced polypeptide chain},
  author={Anfinsen, Christian B and Haber, Edgar and Sela, Michael and White Jr, FH},
  journal={Proceedings of the National Academy of Sciences},
  volume={47},
  number={9},
  pages={1309--1314},
  year={1961}
}

@article{mccammon1977dynamics,
  title={Dynamics of folded proteins},
  author={McCammon, J Andrew and Gelin, Bruce R and Karplus, Martin},
  journal={nature},
  volume={267},
  number={5612},
  pages={585--590},
  year={1977},
  publisher={Nature Publishing Group UK, London}
}

\end{document}